\documentclass[journal,comsoc]{IEEEtran}
\usepackage[T1]{fontenc}

\usepackage{cite}

\ifCLASSINFOpdf
\else
\fi
\usepackage{graphicx}
\usepackage{caption}
\usepackage{subcaption}
\usepackage{enumitem}
\setlist{nosep}
\usepackage{amsfonts}
\usepackage{bbm}
\usepackage{footnote}
\usepackage{colortbl}
\usepackage{amsmath}
\usepackage{empheq}
\usepackage{xcolor}
\usepackage{makecell}
\usepackage{enumitem}
\usepackage{stmaryrd}
\usepackage{array}
\usepackage{amssymb}
\usepackage{booktabs}
\usepackage{multirow}
\usepackage{pifont}         
\newcommand{\xmark}{\ding{55}}%
\newcommand\scalemath[2]{\scalebox{#1}{\mbox{\ensuremath{\displaystyle #2}}}}

\usepackage[cmintegrals]{newtxmath}
\usepackage{lipsum}
\usepackage{xpatch}

\makeatletter
\newinsert\copyrightnote@ins
\count\copyrightnote@ins=1000
\dimen\copyrightnote@ins=8in
\newcommand*\copyrightnote@hook
  {%
    \unvbox\copyrightnote@ins
    \global\let\@makecol\copyrightnote@makecol
  }
\let\copyrightnote@AtBeginDocument\AtBeginDocument
\AtBeginDocument{\let\copyrightnote@AtBeginDocument\@firstofone}
\newcommand*\copyrightnote@firstuse
  {%
    \gdef\copyrightnote@firstuse
      {\global\skip\copyrightnote@ins=\bigskipamount\gdef\copyrightnote@firstuse{}}%
    \global\let\copyrightnote@makecol\@makecol
    \xpatchcmd\@makecol{\unvbox\footins}{\unvbox\footins\copyrightnote@hook}
      {}{\GenericError{}{patching @makecol failed}{}{}}
    \copyrightnote@AtBeginDocument
      {%
        \ifvoid\footins
          \insert\footins{}%
        \else
          \global\skip\copyrightnote@ins=\bigskipamount
        \fi
      }%
  }
\newcommand\copyrightnote[1]
  {%
    \copyrightnote@firstuse
    \copyrightnote@AtBeginDocument
      {%
        \insert\copyrightnote@ins
          {%
            \reset@font\footnotesize
            \interlinepenalty\interfootnotelinepenalty
            \splittopskip\footnotesep
            \splitmaxdepth\dp\strutbox
            \floatingpenalty\@MM
            \hsize\columnwidth
            \@parboxrestore
            \color@begingroup
            \parindent1em
            \noindent
            \hskip1.8em
            \ignorespaces #1\@finalstrut\strutbox
            \color@endgroup
          }%
      }%
  }

\begin{document}
%
\title{Safety-compliant Generative Adversarial Networks 
       for Human Trajectory Forecasting}

\author{Parth~Kothari
        and~Alexandre~Alahi}%

%


\maketitle
\copyrightnote{This work has been submitted to the IEEE for possible publication. Copyright may be transferred without notice, after which this version may no longer be accessible.}

\begin{abstract}
   Human trajectory forecasting in crowds presents the challenges of modelling social interactions and outputting collision-free multimodal distribution. Following the success of Social Generative Adversarial Networks (SGAN), recent works propose various GAN-based designs to better model human motion in crowds. Despite superior performance in reducing distance-based metrics, current networks fail to output socially acceptable trajectories, as evidenced by high collisions in model predictions. To counter this, we introduce SGANv2: an improved safety-compliant SGAN architecture equipped with spatio-temporal interaction modelling and a transformer-based discriminator. The spatio-temporal modelling ability helps to learn the human social interactions better while the transformer-based discriminator design improves temporal sequence modelling. Additionally, SGANv2 utilizes the learned discriminator even at test-time via a collaborative sampling strategy that not only refines the colliding trajectories but also prevents mode collapse, a common phenomenon in GAN training. Through extensive experimentation on multiple real-world and synthetic datasets, we demonstrate the efficacy of SGANv2 to provide socially-compliant multimodal trajectories.
\end{abstract}

%


\IEEEpeerreviewmaketitle

\section{Introduction}
\label{sec:intro}

Forecasting the motion of pedestrians in crowds is essential for autonomous systems like self-driving cars and social robots that will potentially co-exist with humans. To successfully predict how humans navigate in crowds, a forecasting model needs to tackle three crucial challenges: \\ 
(1) \textbf{Modelling social interactions}: the model should learn how the trajectory of one person affects another person; \\
(2) \textbf{Physically acceptable outputs}: the model predictions should be physically acceptable, \textit{i.e.}, not undergo collisions; \\
(3) \textbf{Multimodality}: given the history, the model needs to be able to output all futures without missing any mode.


\begin{figure}
\centering
\begin{subfigure}[h]{0.47\textwidth}
    \includegraphics[width=\textwidth]{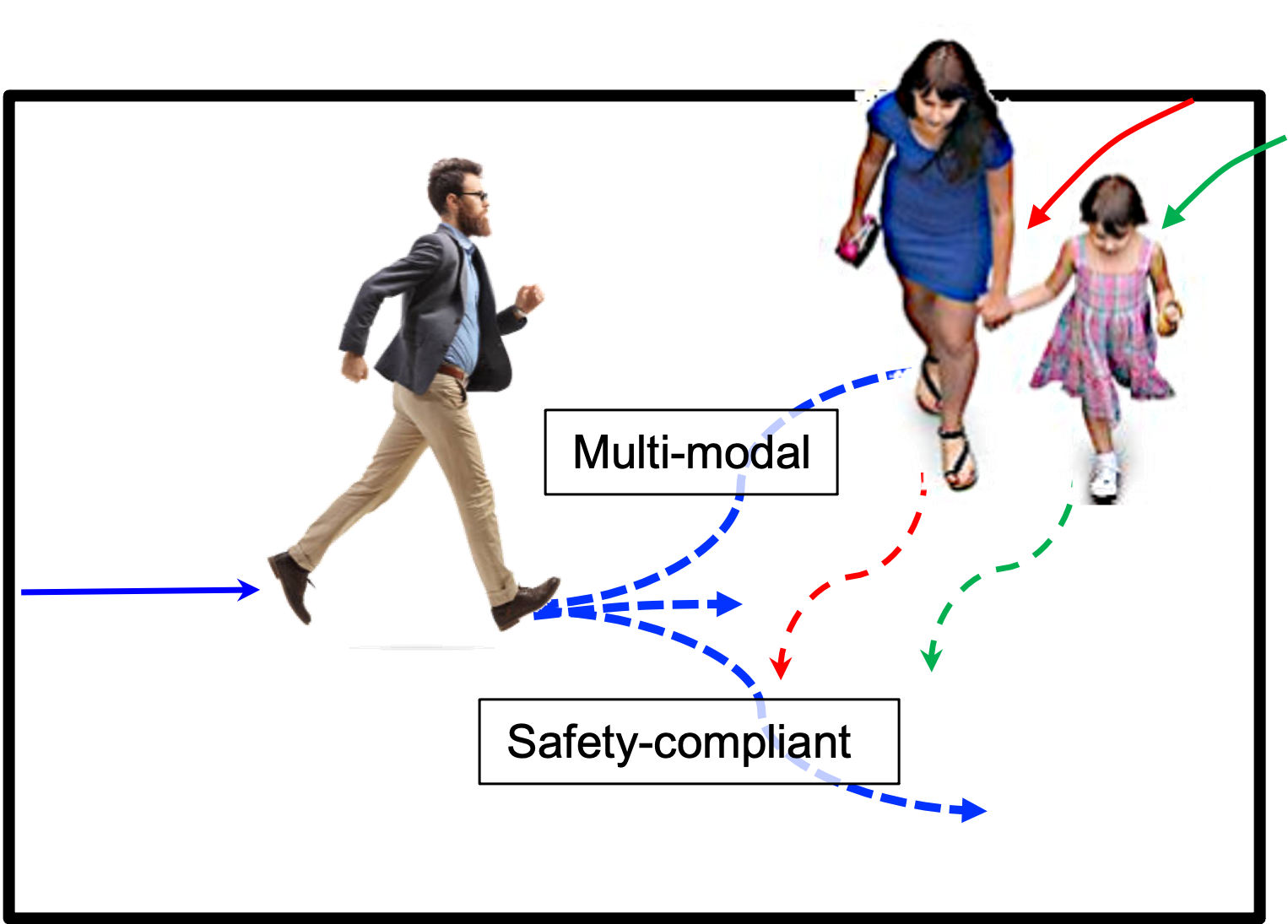}
\end{subfigure}
\setlength{\belowcaptionskip}{-12pt}
\caption{Given the history (solid), a forecasting model needs to account for social rules of human motion when predicting collision-free multimodal futures (dash). SGANv2 learns social interactions using spatio-temporal interaction modelling and refines unsafe outputs via collaborative sampling strategy.}
\label{fig:pull_traj1}
\end{figure}



The objective of multi-modal trajectory forecasting is to learn a generative model over future trajectories. Generative adversarial networks (GANs) \cite{Goodfellow2014GenerativeAN} are a popular choice of generative models for trajectory forecasting as they can effectively capture all possible future modes by mapping samples from a given noise distribution to samples in real data distribution. Gupta \textit{et al.}~\cite{Gupta2018SocialGS} proposed Social GAN (SGAN), GANs with social mechanisms, to learn human interactions and output multimodal trajectories. Following the success of SGAN, recent works \cite{Kosaraju2019SocialBiGATMT, Sadeghian2018SoPhieAA, Zhao2019MultiAgentTF, Amirian2019SocialWL} have proposed improved GAN architectures to better model human interactions in crowds. Indeed, these designs have been successful in reducing the distance-based metrics on real-world datasets \cite{Kosaraju2019SocialBiGATMT}. However, we discover that they fail to model social interactions \textit{i.e.}, the models output colliding trajectories. 




\newcolumntype{?}{!{\vrule width 1pt}}
\renewcommand\theadalign{bc}
\renewcommand\theadfont{\bfseries}
\renewcommand\theadgape{\Gape[4pt]}
\renewcommand\cellgape{\Gape[4pt]}

\begin{table*}[t!]
  \centering
  \resizebox{0.95\textwidth}{!}{\begin{tabular}{lccccccc}
    \toprule
    Method & Generative Model & \makecell{Spatio-temporal \\ Interaction Modelling \\ in Generator}  & \makecell{Multimodal} & \makecell{Spatio-temporal \\ Interaction Modelling \\ in Discriminator} & \makecell{Discriminator \\ Design} & \makecell{Test-time \\ Refinement} \\
    \midrule
    S-LSTM \cite{Alahi2016SocialLH} & -- & \checkmark & \xmark & -- & -- & \xmark \\
    DESIRE \cite{Lee2017DESIREDF}  & VAE & \checkmark & \checkmark & --  & -- & \checkmark    \\
    Trajectron \cite{Ivanovic2018TheTP}  & VAE & \checkmark & \checkmark & --  & -- & \xmark  \\
    SGAN \cite{Gupta2018SocialGS} & GAN & \xmark & \checkmark & \xmark & RNN & \xmark \\
    S-BiGAT \cite{Kosaraju2019SocialBiGATMT} & GAN & \checkmark & \checkmark & \xmark & RNN & \xmark  \\
    SGANv2 [Ours] & GAN & \checkmark & \checkmark & \checkmark & Transformer & \checkmark  \\
    \bottomrule
  \end{tabular}}
  \centering
  \caption{High-level comparison of proposed architecture against selected common generative model-based forecasting models.}
  \label{compare}
\end{table*}

The failure to output collision-free trajectories can be attributed to the fact that the current discriminator designs do not fully model human-human interactions; hence they are incapable of differentiating real trajectory data from fake data. \textit{Only when the discriminator is capable of differentiating real data from fake data, can the supervised signal from it be meaningful to teach the generator}. To tackle this issue, we propose two architectural changes to the SGAN design: (1) Spatio-temporal interaction modelling to better discriminate between real and generated trajectories. (2) A transformer-based discriminator design to strengthen the sequence modelling capability and better guide the generator training. Equipped with these structural changes, our proposed architecture \textit{SGANv2}, learns to better model the underlying etiquette of human motion as evidenced by reduced collisions.



To further reduce the prediction collisions, SGANv2 leverages the trained discriminator even at test time. In particular, we perform collaborative sampling \cite{Liu2019CollaborativeGS} between the generator and discriminator at test-time to guide the unsafe trajectories sampled from the generator. Additionally, we empirically demonstrate that collaborative sampling not only helps to refine trajectories but also has the potential to prevent mode collapse, a phenomenon where the generator fails to capture all modes in the output distribution.



We empirically validate the efficacy of SGANv2 in outputting socially compliant predictions on both synthetic and real-world trajectory datasets. First, we shed light on the shortcomings of the current metric commonly used to measure the multimodal performance, namely Top-20 ADE/FDE \cite{Gupta2018SocialGS}. Specifically, we demonstrate that a simple predictor that outputs uniformly spaced predictions performs at par with the state-of-the-art methods when evaluated using only Top-20 ADE/FDE. To counter this limitation, we propose an alternate evaluation scheme to better measure the socially-compliant multimodal performance of a model. We demonstrate that SGANv2 outperforms competitive baselines on both synthetic and real-world trajectory datasets under the new evaluation scheme. Finally, we demonstrate the ability of collaborative sampling to prevent mode collapse on the recently released Forking Paths \cite{liang2020garden} dataset. Our main contributions are:
\begin{enumerate}[topsep=0pt]
\itemsep0em
\item We propose SGANv2, an improved SGAN architecture that incorporates spatio-temporal interaction modelling in both the generator and the discriminator. Moreover, our transformer-based discriminator better guides the learning process of the generator.
\item We demonstrate the efficacy of collaborative sampling between the generator and discriminator at test-time to reduce prediction collisions and prevent mode collapse in trajectory forecasting.
\end{enumerate}

\begin{figure*}[t!]
\centering
\begin{subfigure}[t]{\textwidth}
    \includegraphics[width=0.98\textwidth]{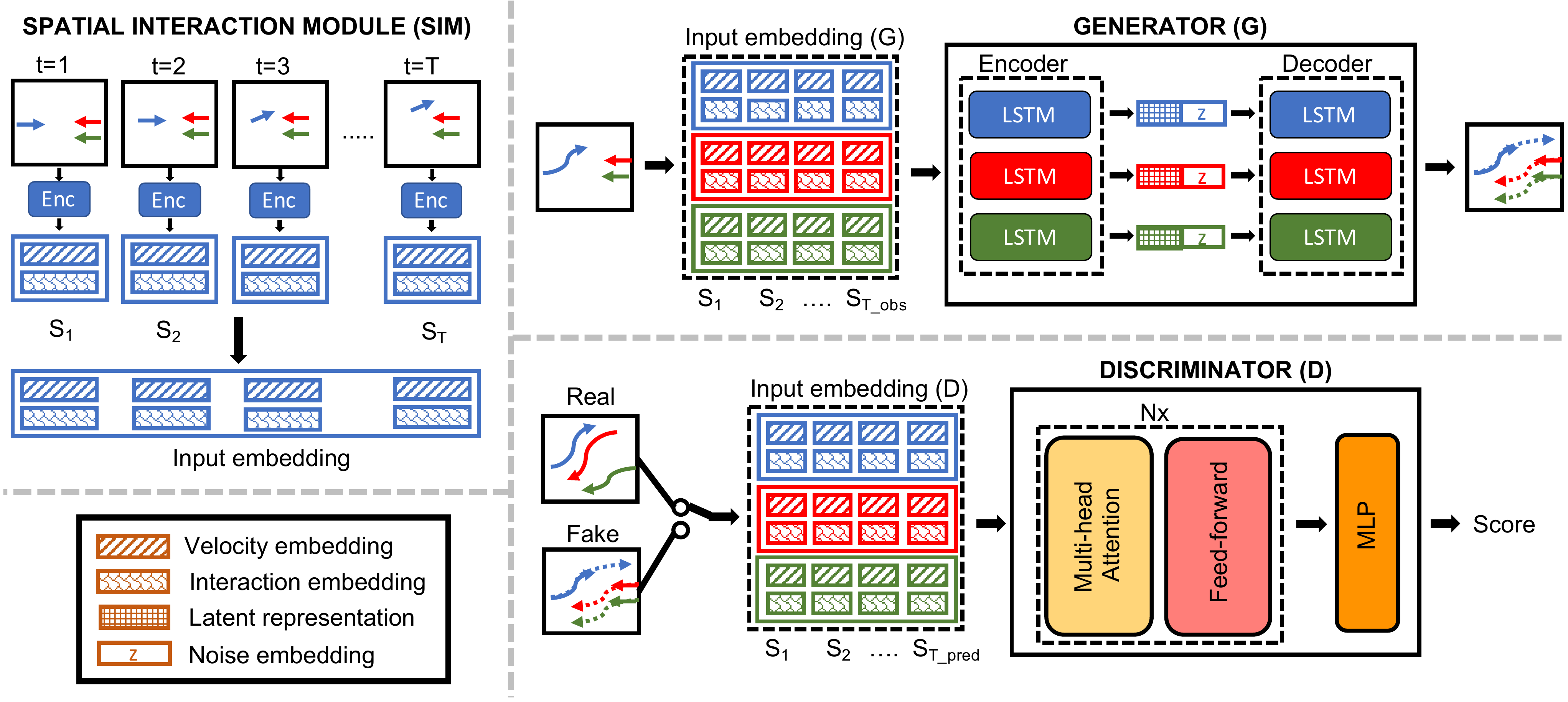}
\end{subfigure}
\caption{Our proposed SGANv2 model: Our model consists of three main parts: the spatial interaction embedding module (SIM), the generator (G) and the discriminator (D). At each time-step, for each pedestrian, the SIM outputs the motion embedding and the spatial interaction embedding. G encodes the input embedding sequence using the encoder LSTM to obtain the latent representation. The latent representation along with the sampled noise vector $z$ is used to generate multimodal predictions using the decoder LSTM. D inputs the stacked embedding sequence of real (or fake) trajectories, encodes it using the transformer architecture to obtain the real (or fake) score.}
\label{fig:sys_fig}
\end{figure*}

\section{Related Work}
Human trajectory forecasting in crowds has been an active area of research \cite{SocialForce, Alahi2016SocialLH, Li2020SocialWaGDATIT, Huang2019STGATMS, Mohamed2020SocialSTGCNNAS, Zhu2019StarNetPT, Giuliari2020TransformerNF, Yu2020SpatioTemporalGT, Su2022TrajectoryFB, Zhang2019SRLSTMSR, Kothari2021InterpretableSA, KothariAdversarialLF, Liu2021SocialNC, Daniel2021PECNetAD, saadatnejad_sattack, Liu2022CausalMotionRepresentations, KothariCoRL2022, Xu2022GroupNetMH, Xu2022AdaptiveTP} for various applications like autonomous systems \cite{WaymoSafety, UberSafety, Chen2019CrowdRobotIC, Rasouli2020AutonomousVT} and advanced surveillance \cite{Mehran2009AbnormalCB}. In this section, we review model designs that learn social interactions and output socially compliant multimodal outputs. Table~\ref{compare} provides a high-level overview of how SGANv2 architecture differs from selected generative model-based designs.


\textbf{Spatio-temporal interaction modelling.} The seminal work of Social LSTM \cite{Alahi2016SocialLH} proposed to learn spatial interactions in a data-driven manner with a novel social pooling layer. Following the success of Social LSTM, various designs of data-driven interaction modules have been proposed \cite{Pfeiffer2017ADM, Shi2019PedestrianTP, Bisagno2018GroupLG, Gupta2018SocialGS, Zhang2019SRLSTMSR, Zhu2019StarNetPT, Ivanovic2018TheTP, Liang2019PeekingIT, Tordeux2019PredictionOP, Ma2016AnAI, Hasan2018MXLSTMMT, Mohamed2020SocialSTGCNNAS, Li2020EvolveGraphHM, Yu2020SpatioTemporalGT, Xu2022GroupNetMH} to effectively model interactions in crowds. For a detailed taxonomy on the designs of interaction modules, one can refer to Kothari \textit{et al.} \cite{Kothari2020HumanTF}. In this work, we highlight the importance of modelling both the spatial and temporal nature of social interactions.

Architectures that model dynamics of entities in spatio-temporal tasks have been well-studied. Structural-RNN \cite{Jain2016StructuralRNNDL}, a specialized RNN design, proposed to model dynamics in spatio-temporal tasks like human-object interaction and driver maneuver anticipation. Specific to motion forecasting, several works consider the temporal evolution of spatial human interactions using recurrent mechanisms \cite{Vemula2017SocialAM, Huang2019STGATMS, Li2020EvolveGraphHM}, graph convolutional networks \cite{Mohamed2020SocialSTGCNNAS, Sun2020RecursiveSB} as well as transformers \cite{Yu2020SpatioTemporalGT}. However, many recent works advocated performing spatial interaction modelling only at the end of observation \cite{Gupta2018SocialGS, Kosaraju2019SocialBiGATMT}, as this strategy did not impact the distance-based metrics and saved computational time. In this work, we study the importance of spatio-temporal interaction modelling from the perspective of reducing the collisions in model outputs.




\textbf{Multimodal forecasting.} Neural networks trained using $L_2$ loss are condemned to output the average of all possible outcomes. To tackle this, one line of work proposes $L_2$ loss variants \cite{GuzmnRivera2012MultipleCL, Rupprecht2016LearningIA, Makansi2019OvercomingLO, Huang2019STGATMS} capable of handling multiple hypotheses. However, these variants fail to penalize low quality predictions, \textit{e.g.}, samples that are far away from the ground truth and undergo collisions. Thus, training using these variants can result in high diversity but low quality predictions. 


Another line of work utilizes generative models \cite{Lee2017DESIREDF, Ivanovic2018TheTP, Gupta2018SocialGS, Amirian2019SocialWL, Kosaraju2019SocialBiGATMT, Huang2021STIGANMP}, with Variational Autoencoders (VAEs) and Generative Adversarial networks (GANs) being the most popular, to model future trajectory distribution. VAE models in trajectory forecasting \cite{Lee2017DESIREDF, Ivanovic2018TheTP} employ a loss objective based on different variants of the euclidean distance. Such a formulation leads to low quality samples especially when the predictions are uncertain \cite{Dosovitskiy2016GeneratingIW}. In contrast, the discriminator of the GAN framework acts as a learned loss function that naturally penalizes the low quality samples under the adversarial training objective \textit{i.e.}, penalty is incurred on the generator if a sample does not look real \cite{Goodfellow2014GenerativeAN}. Thus, we choose GANs as our generative model as they can effectively produce diverse and high-quality modes by transforming samples from a noise distribution to samples in the real data. 

\textbf{GANs in trajectory forecasting.} SGAN \cite{Gupta2018SocialGS} used an LSTM encoder-decoder with social mechanisms within the GAN framework \cite{goodfellow_generative_2014} to perform multimodal forecasting. Following the success of SGAN, various GAN-based architectures have been proposed to better model multimodality in crowds \cite{Li2019WhichWA, Kosaraju2019SocialBiGATMT, Amirian2019SocialWL} as well as on roads \cite{Roy2019VehicleTP, Jin2022AGS}. Yuke Li \cite{Li2019WhichWA} proposed to infer the latent decisions of the agents to model multimodality. Kosaraju \textit{et. al.} \cite{Kosaraju2019SocialBiGATMT} proposed to introduce two discriminators: a local discriminator for the local pedestrian trajectories, similar to \cite{Amirian2019SocialWL, Gupta2018SocialGS}, and a global discriminator that accounted for the spatial interactions. All these works exhibit two common design choices: (1) they do not perform spatio-temporal interaction modelling within the discriminator, (2) they utilize a recurrent LSTM-based discriminator.

It is crucial to equip the discriminator with the ability to model spatio-temporal interactions. Therefore, SGANv2 performs \textit{spatio-temporal interaction modelling} within the discriminator, along with the generator. Transformers \cite{Vaswani2017AttentionIA} have been shown to outperform RNNs in almost all sequence modelling tasks, including trajectory forecasting \cite{Giuliari2020TransformerNF, Yu2020ImprovedOI}. Therefore, we design our discriminator using the transformer and demonstrate that it better guides the generator training. Giuliari \textit{et al. }\cite{Giuliari2020TransformerNF} do not take into account social interactions leading to high collisions in the outputs. 
The spatio-temporal transformer design of STAR \cite{Yu2020SpatioTemporalGT} is most closely related to the design of our discriminator. However, as discussed above, their $L_2$ loss training objective can fail to effectively model multimodality. Further, in contrast to previous transformer and GAN-based works, SGANv2 performs test-time refinement that leads to further collision reduction, discussed next. 


\textbf{Test-time Refinement.} This refers to the task of refining model predictions at test-time. Lee \textit{et al.} \cite{Lee2017DESIREDF} propose an inverse optimal control based module to refine the predicted trajectories. Sun \textit{et al.} \cite{Sun2020ReciprocalLN} refine trajectories using a reciprocal network that reconstructs input trajectories given the predictions. However, they rely on the strong assumption that both forward and backward trajectories follow identical rules of human motion. We propose to refine trajectories by performing collaborative sampling between the trained generator and discriminator \cite{Liu2019CollaborativeGS}. This technique provides theoretical guarantees with respect to moving the generator distribution closer to real distribution.

\textbf{Mode Collapse.} This is the phenomenon where the generator distribution fails to capture all modes of target distribution. SGAN collapses to a single mode of behavior. Social Ways \cite{Amirian2019DataDrivenCS} utilizes InfoGAN that overcomes this issue albeit on a toy dataset. We empirically show that the collaborative sampling technique in SGANv2 overcomes mode collapse on the more-diverse Forking Path dataset \cite{liang2020garden}.




\section{Method}
Modelling human trajectories using generative adversarial networks (GANs) has the potential to learn the underlying etiquette of human motion and output realistic multimodal predictions. Indeed, recent GAN-based trajectory forecasting models have been successful in reducing distance-based metrics, however they suffer from high prediction collisions. In this section, we present SGANv2, an improvement over the SGAN architecture to output safety-compliant predictions. On a high level, we propose three structural changes: (1) \textit{Spatio-temporal interaction modelling} within the discriminator and generator to better understand social interactions, (2) \textit{Transformer-based discriminator} to better guide the generator, (3) \textit{Collaborative sampling mechanism} between the generator and discriminator to refine the colliding trajectories at test-time. Our proposed changes are generic and can be employed on top of any existing GAN-based architecture.

\subsection{Problem Definition}
Given a scene, we receive as input the trajectories of all people within the scene denoted by $\mathbf{X} = \{X_1, X_2, ... X_n\}$, where $n$ is the number of people in the scene. The trajectory of a person $i$, is defined as $X_i = (x_i^t,y_i^t)$, for time $t=1,2...T_{obs}$ and the future ground-truth trajectory is defined as $Y_i = (x_i^t,y_i^t)$ for time $t=T_{obs}+1,...T_{pred}$. The objective is to accurately and simultaneously forecast the future trajectories of all people $\mathbf{\hat{Y}}=\hat{Y}_1,\hat{Y}_2...\hat{Y}_n$, where $\hat{Y}_i$ is used to denote the predicted trajectory of person $i$. The velocity of a pedestrian $i$ at time-step $t$ is denoted by ${v}^{t}_{i}$.

\subsection{Generative Adversarial Networks}
GANs consist of two neural networks, namely the generator $G$ and the discriminator $D$, which are trained together in tandem. The objective of $D$ is to correctly identify whether a sample belongs to the real data distribution or is generated by the generator. The objective of $G$ is to produce realistic samples which can fool the discriminator. $G$ takes as input a noise vector ${z}$ sampled from a given noise distribution ${p_z}$ and transforms it into a real looking sample ${G(z)}.$ $D$ outputs a probability score indicating whether a sample comes from the generator distribution ${p_g}$ or the real data distribution ${p_r}$. Training GANs is essentially a minimax game between the generator and the discriminator: 
\begin{equation} \label{eq:minimax}
    \min_G \max_D \mathbb{E}_{x \sim p_r} [ \log(D(x)) ] + \mathbb{E}_{z \sim p_z} [1 - \log(D(G(z)))].
\end{equation}

\subsection{Interaction Modelling Designs}
Modelling social interactions is the key to outputting safe and accurate future trajectories. In this work, we argue that current works do not model interactions between agents sufficiently within both the generator and discriminator leading to large number of prediction collisions. Here, we differentiate between the notion of performing \textit{spatial interaction modelling} and performing \textit{spatio-temporal interactions modelling}. On one hand, an architectural design is said to perform \textit{spatial interaction modelling} if it models the interaction between pedestrians at a \textbf{single time-step only}. For instance, SGAN performs spatial interaction modelling within the generator as it encodes the neighbourhood information only once, at the end of the observation. On the other hand, an architectural design is said to perform \textit{spatio-temporal interaction modelling} if it performs the spatial interaction modelling at \textbf{every} time-step (from $t=1$ to $t=T_{pred}$) and the temporal evolution of the interactions are captured using any sequence encoding mechanism, \textit{e.g.}, an LSTM or a Transformer. We empirically demonstrate that spatio-temporal interactions modelling within both the generator and the discriminator are essential to output safer trajectories.

\subsection{SGANv2}

We now describe our proposed model design in detail (see Fig.~\ref{fig:sys_fig}). Our architecture consists of three key components: the Spatial Interaction embedding Module (SIM), the Generator (G), and the Discriminator (D). SIM is responsible for spatial interaction modelling while the G and D perform temporal modelling. Thus, \textit{G and D in congregation with SIM perform spatio-temporal interaction modelling (STIM)}. In particular, SIM performs motion embedding and spatial interaction embedding for each pedestrian at each time-step. G encodes the embedded sequence through time and outputs multimodal predictions using an LSTM encoder-decoder framework. D, modelled using a transformer \cite{Vaswani2017AttentionIA}, inputs the entire sequence comprising the observed trajectory $\mathbf{X}$ and the future prediction $\mathbf{\hat{Y}}$ (or ground-truth $\mathbf{Y}$), and classifies it as real/fake. 

\textbf{Spatial Interaction Embedding Module.} One important characteristic that differentiates human motion forecasting from other sequence prediction tasks is the presence of social interactions: the trajectory of a person is affected by other people in their vicinity. SIM performs the task of encoding human motion and human-human interactions in the spatial domain at a particular time-step. 
We embed the velocity ${v}^{t}$ of pedestrian $i$ at time $t$ using a single layer MLP to get the motion embedding vector ${e}^{t}_{i}$ given as:
\begin{equation}
    e^t_{i} = \phi(v^t_i; W_{emb}),
\end{equation}
where $\phi$ is the embedding function with weights $W_{emb}$.

The design of SIM is flexible and it can utilize any spatial interaction module proposed in literature \cite{Kothari2020HumanTF, Kosaraju2019SocialBiGATMT}. It embeds the spatial configuration of the scene and outputs the interaction embedding ${p}^t_{i}$ for pedestrian $i$ at time-step $t$. We then concatenate the motion embedding with the spatial interaction embedding, \textit{i.e.}, ${s}^t_{i} = [{e}^t_{i}; {p}^{t}_{i}]$, and provide the concatenated embedding ${s}^t_{i}$ to the G (or the D). The input embedding is constructed using the ground-truth observations from $[1, T_{obs}]$, and generator predictions from $[T_{obs} + 1, T_{pred}]$. 

\textbf{Generator.} Within the generator, the encoder LSTM encodes the input embedding sequence provided by the SIM. The encoder LSTM helps to model the temporal evolution of spatial interactions in the form of the following recurrence:
\begin{equation} \label{eq:LSTM_main}
    h^t_{i} = LSTM_{enc}(h^{t-1}_i, s^t_{i}; W_{\mathrm{encoder}}),
\end{equation}
where ${h}^t_{i}$ denotes the hidden state of pedestrian $i$ at time $t$, $W_{\mathrm{encoder}}$ are the weights of encoder LSTM that are learned. 

The output of the LSTM encoder for each pedestrian at the end of the observation period represents his/her \textit{observed scene representation}. Similar to SGAN, we utilize this representation to condition our GAN for prediction. In other words, SGANv2 take as input noise $z$ and the observed scene representation to produce future trajectories that are conditioned on the past observations. The decoder hidden-state of each pedestrian is initialized with the final hidden-state of the encoder LSTM. The input noise $z$ is concatenated with the inputs of the decoder LSTM, resulting in the following recurrence for the decoder LSTM:
\begin{equation} \label{eq:LSTM_main_dec}
    h^t_{i} = LSTM_{dec}(h^{t-1}_i, [s^t_{i}; z_{i}]; W_{\mathrm{decoder}}),
\end{equation}
where $W_{\mathrm{decoder}}$ are the weights of decoder LSTM.

The decoder hidden-state at time-step $t$ of pedestrian $i$ is then used to predict the next velocity at time-step $t + 1$. Similar to Alahi \textit{et al.}\cite{Alahi2016SocialLH}, we model the next velocity as a bivariate Gaussian distribution parametrized by the mean $\mu^{t+1} = (\mu_x,\mu_y)^{t+1}$, standard deviation $\sigma^{t+1} = (\sigma_x,\sigma_y)^{t+1}$ and correlation coefficient $\rho^{t+1}$:

\begin{equation}
    [\mu^{t}, \sigma^{t}, \rho^{t}] = \phi_{dec}(h_i^{t-1}, W_{\mathrm{norm}}),
\end{equation}
where $\phi_{dec}$ is an MLP and $W_{norm}$ is learned.

\textbf{Discriminator.} 
The social interactions between humans evolve with time. Therefore, we design our discriminator to perform spatio-temporal interaction modelling. Also, in recent times, transformers \cite{Vaswani2017AttentionIA} have become the de-facto model for modelling temporal sequences, replacing recurrent architectures \cite{Giuliari2020TransformerNF, Yu2020SpatioTemporalGT}. Therefore, we design the discriminator as a transformer to perform the temporal sequence modelling of the output provided by SIM. 

The discriminator takes as input $\textit{Traj}_{\textit{real}} = [\mathbf{X}, \mathbf{Y}]$ or $\textit{Traj}_{\textit{fake}} = [\mathbf{X}, \mathbf{\hat{Y}}]$ and classifies them as real/fake. The discriminator has its own SIM, which provides the spatial interaction embedding $s^{t}_{i}$ for each pedestrian $i$ at each time-step $t$ in the input sequence. Instead of passing $s^{t}_{i}$ through an LSTM (similar to the generator), we stack these embedded vectors together to form an embedded sequence $S_{i}$ for each pedestrian $i$ (similar to an embedded sequence obtained after embedding word tokens in the field of natural language \cite{Vaswani2017AttentionIA}):
\begin{equation}
  S_{i}  = [s^{1}_{i}; s^{2}_{i}; \ldots s^{T_{pred}}_{i}].
\end{equation}
This sequence $S_{i}$ is given as input to the encoder of the transformer proposed in \cite{Vaswani2017AttentionIA}. The ability of transformers to capture the temporal correlations within the spatial interaction embedding lies mainly in its self-attention module. Within the attention module, each element of the sequence $S_{i}$ is decomposed into query (Q), key (K) and value (V). The matrix of outputs is computed using the following equation \cite{Vaswani2017AttentionIA}:
\begin{equation}
    \text{Attention}(Q, K, V) = \text{softmax}\left(\frac{QK^{T}}{\sqrt{d_k}}\right)V,
\end{equation}
where $d_k$ is the dimension of the SIM embedding $s^{t}_{i}$. The output of the attention layer is normalized and passed through a feedforward layer to obtain the latent representation of the input sequence, denoted by $R_{i}$: 
\begin{equation}
    R_{i} = \max (0, A_{i} * W1 + b1) * W2 + b2,
\end{equation}
where the weights $W1, W2, b1, b2$ are learned, $*$ represents matrix multiplication and $A_{i}$ denotes the normalized representation of the output of the attention module. We utilize the last element of $R_{i}$, as the representation of the input sequence. This embedding gets scored using an MLP $\phi_{d}$ to determine if the sequence is real or fake.

\subsection{Training}
As mentioned earlier, SGANv2 is a conditional GAN model. It takes as input noise vector $z$, sampled from $\mathcal{N}(0, 1)$, and outputs future trajectories $\mathbf{\hat{Y}}$ conditioned on the past observations $\mathbf{X}$. We found the least-square training objective \cite{Mao2017LeastSG} to be effective in training SGANv2:
\begin{align} \label{eq:ls_minimax}
&\scalemath{0.8}{\min_{G} \mathcal{L}(G) =  \frac{1}{2} \mathbb{E}_{z \sim p_z} [(D(X, G(X, z)) - 1)^{2}]}, \\
&\scalemath{0.8}{
\min_{D} \mathcal{L}(D) =  \frac{1}{2} \mathbb{E}_{x \sim p_r} [ (D(X, Y) - 1)^{2} ] + \frac{1}{2} \mathbb{E}_{z \sim p_z} [(D(X, G(X, z)))^{2}]}.
\end{align}


Additionally, we utilize the variety loss \cite{Gupta2018SocialGS} to further encourage the network to produce diverse samples. For each scene, we generate $k$ output predictions by randomly sampling $z$ and penalize the prediction closest to the ground-truth based on L2 distance.
\begin{equation}
    \mathcal{L}_{variety} = \min_{k} \| Y - G(X, z)^{(k)} \|_{2}^{2}.
\end{equation}


Following the strategy in \cite{Kothari2020HumanTF}, the generator predicts only the trajectory of the pedestrian of interest in each scene and uses the ground-truth future of neighbours during training. During test time, we predict the trajectories of \textit{all} the pedestrians simultaneously in the scene. All the learnable weights are shared between all pedestrians in the scene. 


\begin{figure}
\centering
\begin{subfigure}[h]{0.40\textwidth}
    \includegraphics[width=\textwidth]{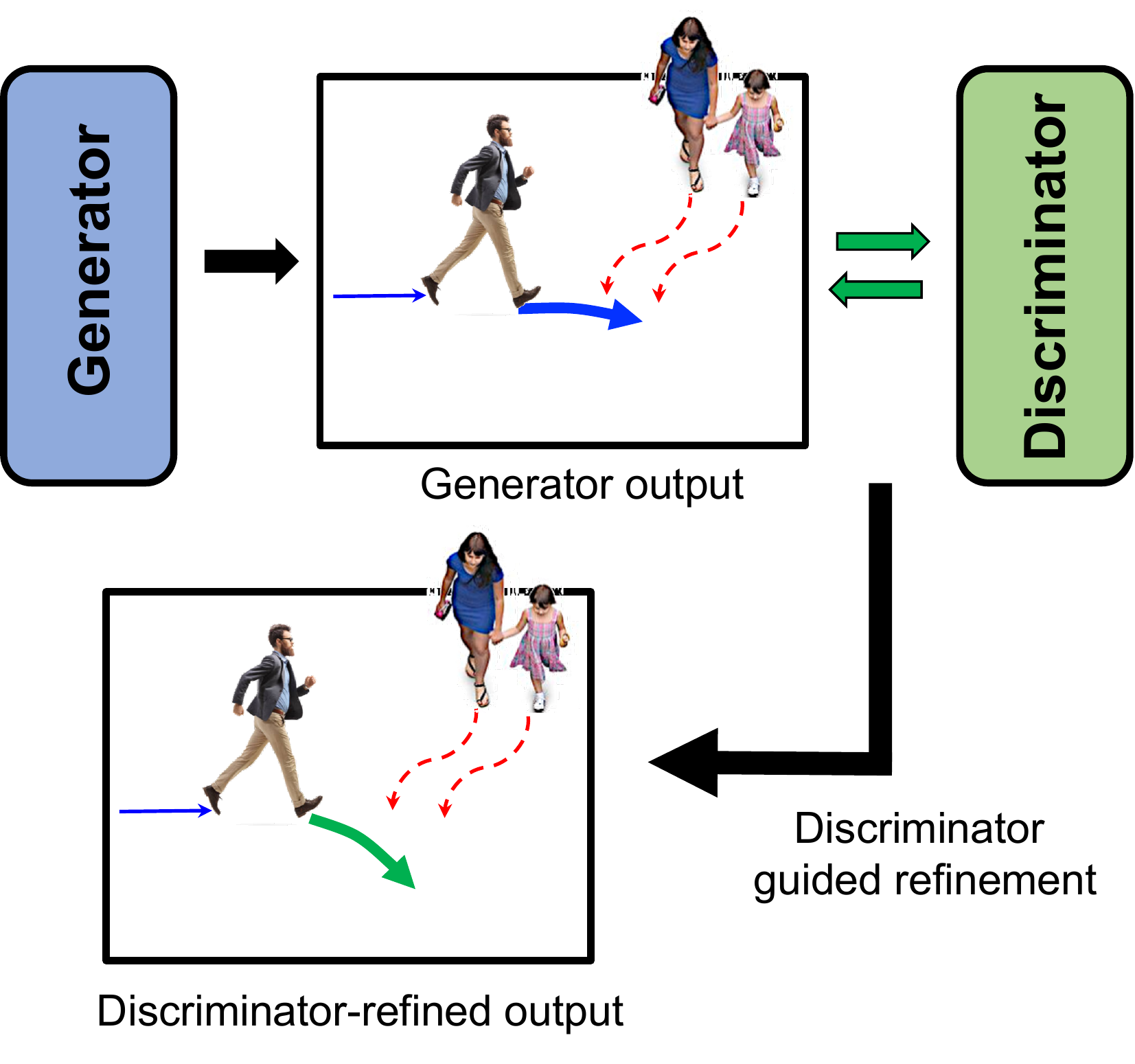}
\end{subfigure}
\setlength{\belowcaptionskip}{-12pt}
\caption{Illustration of trajectory refinement using collaborative sampling. The trained discriminator provides feedback to improve the generated samples during test-time.}
\label{fig:collab}
\end{figure}

\subsection{Collaborative Sampling in GANs}\label{Collab-section}
The common practice in GANs is to sample from the generator and discard the discriminator during test time. However, our trained discriminator has knowledge regarding the social etiquette of human motion. We can utilize this knowledge to refine the \textit{bad} predictions proposed by the generator. We define a prediction as \textit{bad} if the pedestrian of interest undergoes collision in the model prediction. We propose to refine such trajectories by performing collaborative sampling \cite{Liu2019CollaborativeGS} between the generator and discriminator, as demonstrated in Fig.~\ref{fig:collab}.

\begin{table*}[t]
    \centering
    \resizebox{\textwidth}{!}{\begin{tabular}{lccc|ccc|ccc|ccc|ccc}
    \toprule
 \multicolumn{1}{l}{\textbf{Model}} & \multicolumn{3}{c}{\textbf{ETH}} & \multicolumn{3}{c}{\textbf{HOTEL}} & \multicolumn{3}{c}{\textbf{UNIV}} & \multicolumn{3}{c}{\textbf{ZARA1}} &\multicolumn{3}{c}{\textbf{ZARA2}}\\ \rowcolor{gray!15}
 \; & Top-3 & Top-20 & Col
 & Top-3 & Top-20 & Col
 & Top-3 & Top-20 & Col
 & Top-3 & Top-20 & Col
 & Top-3 & Top-20 & Col\\ \midrule

 Transformer$^\dagger$ \cite{Giuliari2020TransformerNF}
  & 1.0/1.9 & 0.6/0.9 & 5.8
  & 0.5/0.9 & 0.3/0.5 & 8.2
  & 2.3/4.2 & 0.8/1.3 & 10.9
  & 0.5/1.0 & 0.3/0.4 & 7.1
  & 0.4/0.8 & \textbf{0.2}/\textbf{0.3} & 11.3 \\ \rowcolor{gray!15}

 STGAT$^\dagger$ \cite{Huang2019STGATMS}
  & \textbf{0.9}/\textbf{1.8} & 0.7/1.2 & 1.7
  & 0.7/1.4 & 0.5/1.0 & 4.2
  & 0.6/1.2 & 0.3/0.7 & 13.9 
  & 0.4/0.9 & \textbf{0.2}/\textbf{0.4} & 3.9 
  & 0.4/0.7 & 0.2/0.4 & 6.9 \\ 
  
 Social-STGCNN$^\dagger$ \cite{Mohamed2020SocialSTGCNNAS}
  & 1.0/1.8 & 0.7/1.2 & 6.7
  & 0.4/0.8 & 0.3/0.6 & 10.4 
  & 0.7/1.3 & 0.5/0.8 & 25.0 
  & 0.5/0.9 & 0.3/0.5 & 12.1 
  & 0.4/0.8 & 0.3/0.5 & 19.4 \\ \rowcolor{gray!15}
  
  Uniform Predictor (UP)
  & 1.1/2.2 & \textbf{0.6}/\textbf{0.9} & 3.3
  & 0.5/0.9 & \textbf{0.2}/\textbf{0.4} & 5.1 
  & 0.6/1.3 & \textbf{0.3}/\textbf{0.6} & 15.7
  & 0.5/1.0 & 0.3/0.6 & 4.7
  & 0.4/0.8 & 0.2/0.4 & 7.5 \\
 
 \midrule

  SGANv2 [Ours]
  & 1.0/1.9 & 0.7/1.2 & \textbf{1.0}
  & \textbf{0.4}/\textbf{0.7} & 0.3/0.5 & \textbf{1.2} 
  & \textbf{0.6}/\textbf{1.3} & 0.5/0.8 & \textbf{8.3} 
  & \textbf{0.4}/\textbf{0.8} & 0.3/0.6 & \textbf{1.3}
  & \textbf{0.3}/\textbf{0.7} & 0.3/0.5 & \textbf{2.2} \\
  

\bottomrule
 
\end{tabular}}
\setlength{\belowcaptionskip}{-10pt}
\caption{Quantitative evaluation of various methods on ETH-UCY. Errors reported are Top-K ADE/FDE (in m) and collision (in \%). Only observing the Top-20 metric (as done by previous work) can lead to incorrect conclusions. We show that a high-entropy uniform predictor is highly competitive with respect to state-of-the-art methods in multimodal forecasting using Top 20 metric. See Table \ref{tab:Real} for a more informative evaluation.}
\label{tab:top20}
\end{table*}

To summarize collaborative sampling for the case of trajectory forecasting, our goal is to refine the generator prediction using gradients from the discriminator \textit{without} updating the parameters of the generator. We leverage the gradient information provided by the discriminator to continuously refine the generator predictions of the pedestrian of interest $i$ through the following iterative update:
\begin{equation} \label{eq:backward}
    \hat{Y}_i^{m+1} = \hat{Y}_i^{m} - \lambda \nabla \mathcal{L}_{G} (\hat{Y}_i^m),
\end{equation}
where $m$ is the iteration number, $\lambda$ is the stepsize, $\mathcal{L}_{G}$ is the loss of the generator in Eq.~\ref{eq:ls_minimax}. The authors demonstrate that the above iteration process theoretically, under mild assumptions, shifts the learned generator distribution towards the real distribution \cite{Liu2019CollaborativeGS}. The trajectories are updated till either the discriminator score goes above a defined threshold or the maximum number of iterations is reached.

\section{Experiments}

\begin{figure}
\centering
\begin{subfigure}[h]{0.19\textwidth}
    \includegraphics[width=\textwidth]{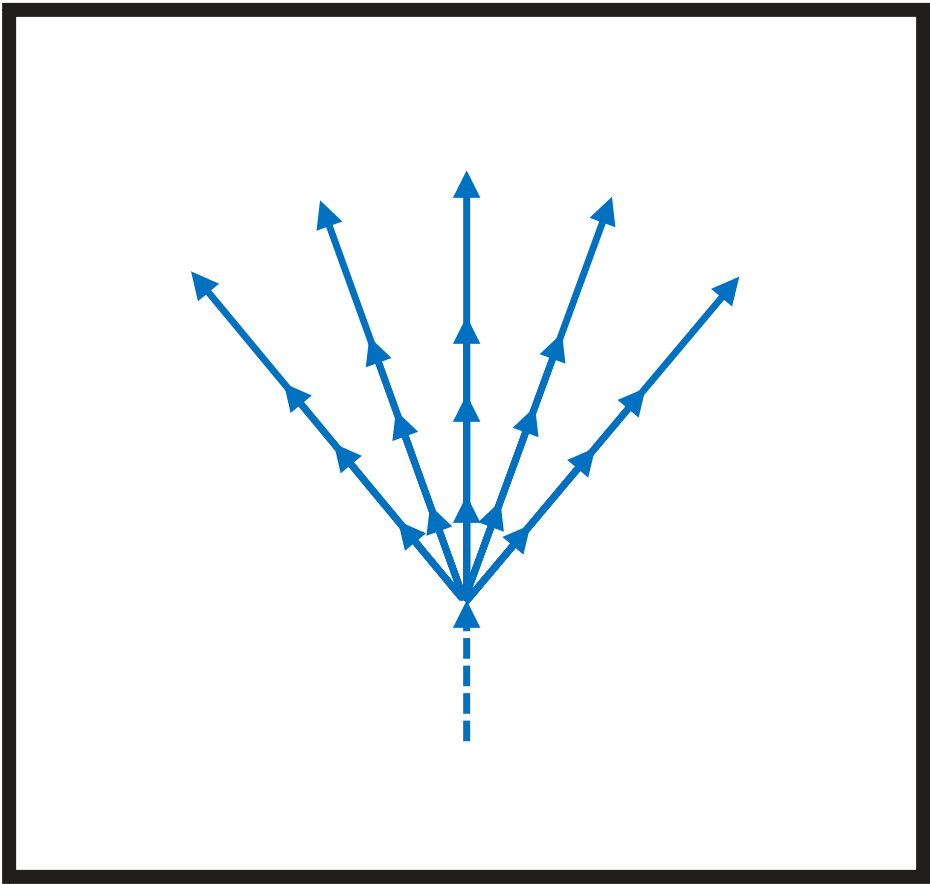}
\end{subfigure}
\setlength{\belowcaptionskip}{-12pt}
\caption{20 uniformly spread predictions (solid) of a handcrafted predictor conditioned on the last observed velocity (dotted).}
\label{fig:sp}
\end{figure}

In this section, we highlight the ability of SGANv2 to output socially-compliant multimodal futures. We evaluate the performance of our architecture against several state-of-the-art methods on the ETH/UCY datasets \cite{Pellegrini2009YoullNW, Lerner2007CrowdsBE} and on the interaction-centric TrajNet++ benchmark \cite{Kothari2020HumanTF}. Additionally, we highlight the potential of collaborative sampling to prevent mode collapse on the Forking Paths \cite{liang2020garden} dataset. We evaluate two variants of our model against various baselines:
\begin{itemize}
\item \textbf{SGANv2 w/o CS}: Our GAN architecture comprising of a transformer-based discriminator that performs spatio-temporal interaction modelling.
\item \textbf{SGANv2}: Our complete GAN architecture in combination with collaborative sampling at test-time.
\end{itemize}



\subsection{Evaluation Metrics}
\begin{enumerate}[itemsep=0.25cm]
    \item \textbf{Top-K Average Displacement Error (ADE)}: Average $l_{2}$ distance between ground truth and closest prediction (out of k samples) over all predicted time steps. 
    \item \textbf{Top-K Final Displacement Error (FDE)}: The distance between the final destination of closest prediction (out of k samples) and the ground truth final destination at the end of the prediction period $T_{pred}$.
    \item \textbf{Prediction collision (Col)} \cite{Kothari2020HumanTF}: The percentage of collision between the primary pedestrian and the neighbors in the \textit{forecasted future} scene.
\end{enumerate}

\subsection{Limitations of current multimodal evaluation scheme}
Current multimodal forecasting works utilize metrics that measure model performance at the \textit{individual level} such as the top-$k$ ADE/FDE \cite{Gupta2018SocialGS, Huang2019STGATMS}. This metric evaluates the quality of the predicted distribution per pedestrian; and does not measure the interaction between different pedestrians. Further, the value of $k$ is very high (k=20 being most common). Almost all the recent works \cite{Gupta2018SocialGS, Kosaraju2019SocialBiGATMT, Daniel2021PECNetAD, Huang2019STGATMS, Giuliari2020TransformerNF} in human trajectory forecasting utilize the Top-20 ADE/FDE metric \cite{Gupta2018SocialGS} to quantify multimodal performance. We argue that measuring multimodal performance based solely on this metric can be misleading. 

The Top-20 ADE/FDE metric can be easily cheated by predicting a high entropy distribution that covers all the space but is not precise \cite{Eghbalzadeh2017LikelihoodEF}. We empirically validate this claim by comparing state-of-the-art baselines against a simple hand-crafted uniform predictor (\texttt{UP}). \texttt{UP} takes as input the last observed velocity of each pedestrian and outputs 20 uniformly spread trajectories (see Fig~\ref{fig:sp}). \texttt{UP} outputs 20 predictions using the combination of 5 different relative direction profiles $[0, 25, 50, -25, -50]$ (in degrees relative to current direction of motion) and 4 different relative speed profiles $[1, 0.75, 1.25, 0.25]$ (factors of the current speed).

Table~\ref{tab:top20} compares the performance of recent state-of-the-art methods \cite{Giuliari2020TransformerNF, Huang2019STGATMS, Mohamed2020SocialSTGCNNAS} and \texttt{UP} on ETH-UCY datasets. It is apparent that by observing the \textit{Top-20 metric only}, \texttt{UP} seems to perform better (or at par) against the state-of-the-art baselines. If we note the prediction collisions, it is apparent that \texttt{UP} is not a good multimodal predictor. This corroborates our conjecture that a high entropy distribution can easily cheat the Top-20 metric leading to incorrect conclusions. 

\begin{figure*}[t!]
    \centering
    \includegraphics[width=0.98\textwidth]{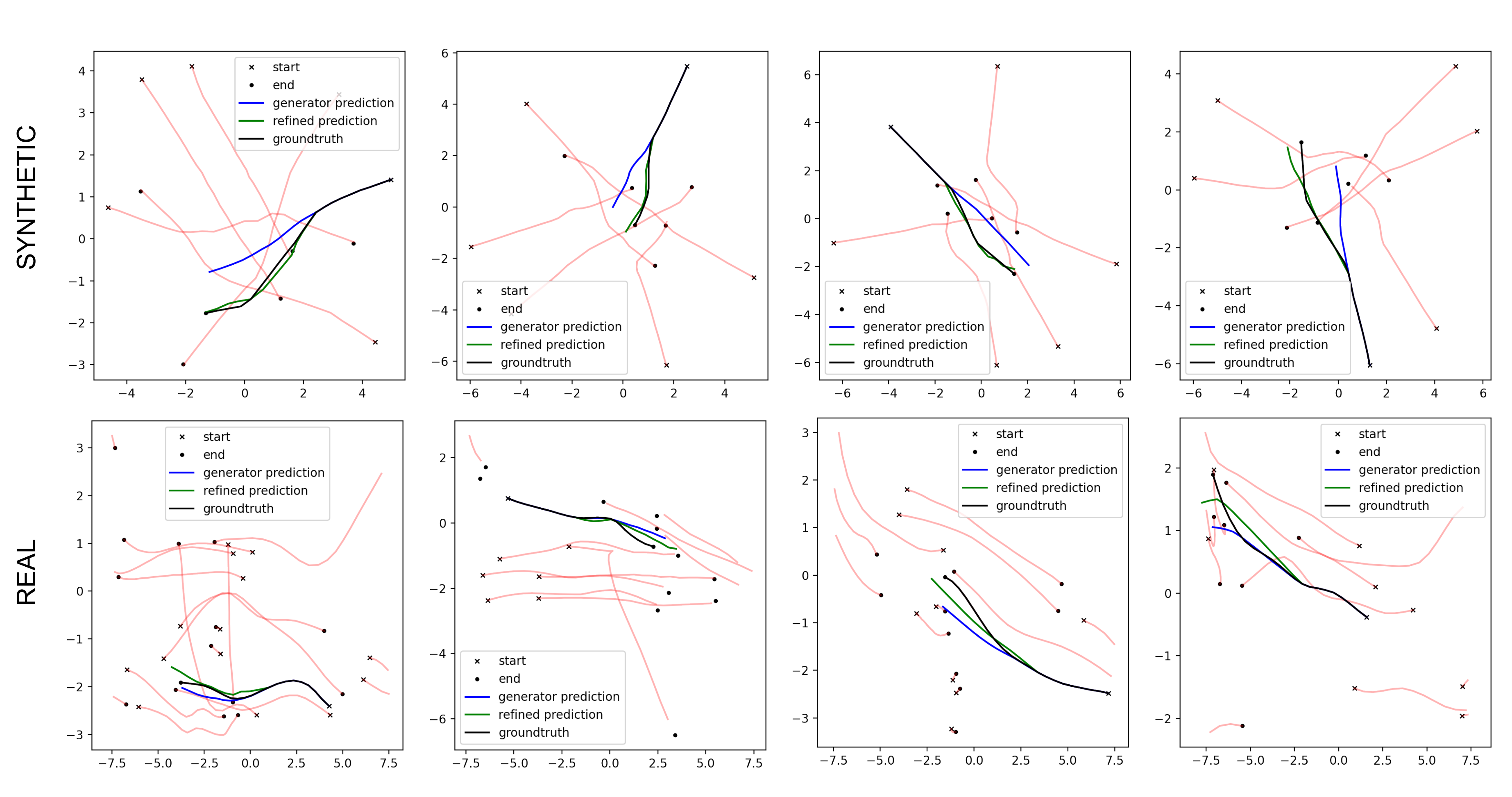}
    \caption{Illustration of collaborative sampling at test-time to reduce model collisions in both TrajNet++ synthetic and real-world datasets. Given a generator prediction of the pedestrian of interest (blue) that undergoes collision with the neighbours (red), our discriminator, equipped with spatio-temporal interaction modelling, provides feedback based on its learned understanding human-human interactions. Consequently, the resulting refined prediction (green) does not undergo collision and in some cases, is closer to the ground-truth (black)}
    \label{fig:collab_sample_viz}
\end{figure*}

\subsection{Multimodal Evaluation Scheme}

To counter the above issues with current multimodal evaluation strategy, we propose to set $k$ to a lower value in our experiments; as a lower $k$ is a better proxy for likelihood estimation for implicit generative models \cite{Eghbalzadeh2017LikelihoodEF}. Specific to our problem, we will demonstrate that when $k$ is low ($k = 3$), the uniform predictor due to a lack of modeling social interactions performs poorly compared to interaction-based baselines \cite{Huang2019STGATMS, Mohamed2020SocialSTGCNNAS}. Further, to measure the interaction-modelling capability, we focus on the percentage of collisions between the primary pedestrian and the neighbors in the \textit{forecasted future} scene.

\subsection{Synthetic Experiments}
We first demonstrate the efficacy of our proposed architectural changes in SGANv2 compared to other generative model designs in the TrajNet++ synthetic setup. We observe that SGANv2 greatly improves upon the Top-3 ADE/FDE metric with a lower collision metric compared to training a model using only variety loss (see Table~\ref{tab:traj_synth}).

\begin{table}[htb!]
    \centering
    \resizebox{0.34\textwidth}{!}{\begin{tabular}{lcc}
    \toprule
   \textbf{Method} & \textbf{Top-3} & \textbf{Col} \\ \midrule
    CV* \cite{Schller2020WhatTC}                 &  0.4/1.0 & 21.1 \\    \rowcolor{gray!15}
    LSTM* \cite{Hochreiter1997LongSM}            &  0.3/0.6 & 19.0 \\ 
    S-LSTM* \cite{Kothari2020HumanTF}            & 0.2/0.5 & 2.2 \\    \rowcolor{gray!15}
    D-LSTM* \cite{Kothari2020HumanTF}            & 0.2/0.5 & 2.2 \\    
    CVAE \cite{Lee2017DESIREDF}                  & 0.2/0.5 & 4.6 \\ \rowcolor{gray!15}
    WTA  \cite{Rupprecht2016LearningIA}  & 0.2/0.4 & 2.4\\   
    SGAN  \cite{Gupta2018SocialGS}  & 0.2/0.4 & 2.8\\ \midrule \rowcolor{gray!15} 
    SGANv2 w/o CS [Ours]     & 0.2/0.4 & 1.9 \\ 
    SGANv2 [Ours]    & \textbf{0.2}/\textbf{0.4} & \textbf{0.6}\\ 
    \bottomrule
    \end{tabular}}
    \caption{Quantitative evaluation on TrajNet++ synthetic dataset. Errors reported are Top-3 ADE/FDE (in m) and collision (in \%). SGANv2 with collaborative sampling greatly reduces the model prediction collisions without compromising on the distance-based metrics. *Unimodal methods}
    \label{tab:traj_synth}
\end{table}

\begin{table*}[t]
    \centering
    \resizebox{0.90\textwidth}{!}{\begin{tabular}{lcccccccccc}
    \toprule
 \multicolumn{1}{l}{\textbf{Model}} & \multicolumn{2}{c}{\textbf{ETH}} & \multicolumn{2}{c}{\textbf{HOTEL}} & \multicolumn{2}{c}{\textbf{UNIV}} & \multicolumn{2}{c}{\textbf{ZARA1}} &\multicolumn{2}{c}{\textbf{ZARA2}}\\ \rowcolor{gray!15}
 \; & Top-3 & Col
 & Top-3 & Col
 & Top-3 & Col
 & Top-3 & Col
 & Top-3 & Col\\ \midrule

 CV* \cite{Schller2020WhatTC} 
 & 1.1/2.3 & 5.3
 & 0.4/0.8 & 7.2 
 & 0.6/1.4 & 20.3 
 & 0.4/1.0 & 6.0 
 & 0.3/0.7 & 9.6 \\ \rowcolor{gray!15}

 LSTM* \cite{Hochreiter1997LongSM} 
  & 1.0/2.1  & 5.8
  & 0.5/0.9  & 6.7 
  & 0.6/1.3  & 20.2 
  & 0.5/1.0  & 5.2 
  & 0.4/0.8  & 9.5  \\ 

 Uniform Predictor 
  & 1.1/2.2 & 3.3
  & 0.5/0.9 & 5.1
  & 0.6/1.3 & 15.7
  & 0.5/1.0 & 4.7
  & 0.4/0.8 & 7.5 \\ \rowcolor{gray!15}

 Transformer$^\dagger$ \cite{Giuliari2020TransformerNF}
  & 1.0/1.9 & 5.8
  & 0.5/0.9 & 8.2
  & 2.3/4.2 & 10.9
  & 0.5/1.0 & 7.1
  & 0.4/0.8 & 11.3 \\ 
 

 S-LSTM* \cite{Alahi2016SocialLH}
  & 1.1/2.1 & 2.2 
  & 0.5/0.9 & 2.5 
  & 0.7/1.5 & 11.8 
  & 0.4/0.9 & 2.7 
  & 0.4/0.8 & 3.7  \\ \rowcolor{gray!15}

 CVAE \cite{Lee2017DESIREDF} 
  & 1.1/2.2 & 2.8
  & 0.4/0.8 & 1.5
  & 0.7/1.5 & 12.6
  & 0.4/0.9 & 2.6
  & 0.4/0.8 & 3.5 \\ 

  WTA \cite{Rupprecht2016LearningIA} 
  & 1.0/1.9 & 2.5
  & 0.4/0.7 & 2.3 
  & 0.6/1.3 & 12.7 
  & 0.4/0.8 & 2.2 
  & 0.3/0.7 & 4.1  \\ \rowcolor{gray!15}

 SGAN \cite{Gupta2018SocialGS} 
  & 1.0/2.0 & 2.2 
  & 0.4/0.7 & 1.7 
  & 0.6/1.3 & 11.8
  & 0.4/0.8 & 2.3
  & 0.3/0.7 & 3.2  \\ 

 STGAT$^\dagger$ \cite{Huang2019STGATMS}
  & \textbf{0.9}/\textbf{1.8} & 1.7
  & 0.7/1.4 & 4.2
  & 0.6/1.2 & 13.9 
  & 0.4/0.9 & 3.9 
  & 0.4/0.7 & 6.9 \\ \rowcolor{gray!15}

 Social-STGCNN$^\dagger$ \cite{Mohamed2020SocialSTGCNNAS}
  & 1.0/1.8 & 6.7
  & 0.4/0.8 & 10.4 
  & 0.7/1.3 & 25.0 
  & 0.5/0.9 & 12.1 
  & 0.4/0.8 & 19.4 \\ 

 S-BiGAT \cite{Kosaraju2019SocialBiGATMT} 
  & 1.0/1.9 & 3.3  
  & 0.4/0.7 & 1.7 
  & 0.6/1.3 & 11.5
  & 0.4/0.8 & 2.2 
  & 0.3/0.7 & 3.3 \\ \rowcolor{gray!15}
 \midrule
 

 SGANv2 w/o CS [Ours]
  & 1.0/1.9 & 1.7
  & 0.4/0.7 & 1.4
  & 0.6/1.3 & 11.5
  & 0.4/0.8 & 2.1
  & 0.3/0.7 & 3.6 \\ 
 
  SGANv2 [Ours]
  & 1.0/1.9 & \textbf{1.0}
  & \textbf{0.4}/\textbf{0.7} & \textbf{1.2} 
  & \textbf{0.6}/\textbf{1.3} & \textbf{8.3} 
  & \textbf{0.4}/\textbf{0.8} & \textbf{1.3}
  & \textbf{0.3}/\textbf{0.7} & \textbf{2.2} \\
\bottomrule
 
\end{tabular}}
\setlength{\belowcaptionskip}{-10pt}
\caption{Quantitative evaluation of our proposed method on ETH/UCY datasets. We observe the trajectories for 8 times steps (3.2 seconds) and show prediction results for the next 12 time steps (4.8 seconds). Errors reported are Top-3 ADE / FDE (in m), Col (in \%). SGANv2 improves in collision metric without compromising on the distance-based metrics. *Unimodal}
\label{tab:Real}
\end{table*}

Next, we utilize collaborative sampling technique to refine trajectories that undergo collision at test-time. The trained discriminator provides feedback to the colliding samples which helps to reduce the collisions. For each colliding prediction, we perform 5 refinement iterations with stepsize 0.01. We observe that this scheme greatly reduces the collision rate by \textbf{$\sim$ 70\%}. The first row of Fig~\ref{fig:collab_sample_viz} illustrates the ability of collaborative sampling to refine predictions in the synthetic scenario.


\begin{table}[htb!]
    \centering
    \resizebox{0.34\textwidth}{!}{\begin{tabular}{lcc}
    \toprule
   \textbf{Method} & \textbf{Top-3} & \textbf{Col} \\ \midrule
    CV* \cite{Schller2020WhatTC}                        &     0.6/1.3 & 10.9  \\    \rowcolor{gray!15}
    LSTM* \cite{Hochreiter1997LongSM}                   &     0.5/1.2 & 9.3  \\ 
    S-LSTM* \cite{Alahi2016SocialLH}                    &     0.5/1.0 & 4.9 \\    \rowcolor{gray!15}
    D-LSTM* \cite{Kothari2020HumanTF}                   &     0.5/1.1 & 3.9 \\
    CVAE \cite{Lee2017DESIREDF}                         &     0.5/1.1 & 3.9 \\ \rowcolor{gray!15}
    WTA  \cite{Rupprecht2016LearningIA}                 &     0.5/1.0 & 3.5 \\   
    SGAN \cite{Gupta2018SocialGS}                       &     0.5/1.0 & 3.5 \\ \rowcolor{gray!15}
    S-NCE \cite{Liu2021SocialNC}                        &     0.5/1.1 & 4.0 \\
    PECNet \cite{Daniel2021PECNetAD}                    &     \textbf{0.4}/\textbf{0.9} & 10.7 \\ \hline \rowcolor{gray!15}
    Uniform Predictor                                   &     0.6/1.2 & 8.4 \\
    Transformer$^\dagger$ \cite{Giuliari2020TransformerNF}.       &     0.7/1.3 & 9.4 \\ \rowcolor{gray!15}
    STGCNN$^\dagger$  \cite{Mohamed2020SocialSTGCNNAS}            &     0.6/1.1 & 12.6 \\ 
    STGAT$^\dagger$ \cite{Huang2019STGATMS}                       &     0.5/1.1 & 5.6 \\ \rowcolor{gray!15}
    S-BiGAT \cite{Kosaraju2019SocialBiGATMT}            &     0.5/1.0 & 3.3 \\ \midrule   
    SGANv2 w/o CS [Ours]    & 0.5/1.0 & 3.1 \\ \rowcolor{gray!15}
    SGANv2  [Ours]       & 0.5/1.0 & \textbf{2.3}\\  
    \bottomrule
    \end{tabular}}
    \caption{Quantitative evaluation of our proposed method on TrajNet++ real-world dataset. Errors reported are Top-3 ADE/FDE (in m) and collision (in \%). SGANv2 in combination with collaborative sampling (CS) improves in collision metric without compromising on the distance-based metrics. *Unimodal}
    \label{tab:traj_real}
\end{table}

\subsection{Real-World Experiments}
Next, we evaluate the performance of our SGANv2 architecture in real-world datasets of ETH/UCY and the TrajNet++ benchmark. For ETH/UCY, we observe the trajectories for 8 times steps (3.2 seconds) and show prediction results for 12 (4.8 seconds) time steps. For TrajNet++, we observe the trajectories for 9 times steps (3.6 seconds) and show prediction results for 12 (4.8 seconds) time steps.

Table~\ref{tab:Real} provides the quantitative evaluation of various baselines and state-of-the-art forecasting methods on the ETH/UCY dataset. We observe that SGANv2 outputs safer predictions in comparison to competitive baselines without compromising on the prediction accuracy. Our Top-3 ADE/FDE are on par with (if not better than) state-of-the-art methods while our collision rate is significantly reduced thanks to spatio-temporal interaction modelling. It is further interesting to note that Trajectory Transformer \cite{Giuliari2020TransformerNF} and the simple uniform predictor (\texttt{UP}) that performed the best on Top-20 ADE/FDE in Table~\ref{tab:top20} are not among the top performing methods when evaluated on the more-strict Top-3 ADE/FDE. Next, we benchmark on the TrajNet++ with interaction-centric scenes with a standardized evaluator that provides a more objective comparison \cite{Kothari2020HumanTF}.



Table~\ref{tab:traj_real} compares SGANv2 against other competitive baselines on TrajNet++ real-world benchmark. The first part of Table~\ref{tab:traj_real} reports simple baselines and the top-3 official submissions on AICrowd made by different works literature \cite{Liu2021SocialNC, Daniel2021PECNetAD, Kothari2020HumanTF}. SGANv2 performs at par with the top-ranked PECNet \cite{Daniel2021PECNetAD} on the Top-3 evaluation while having \textbf{3x} lower collisions demonstrating that spatio-temporal interaction modelling is key to outputting safer trajectories \footnote{PECNet performs spatial interaction modelling once at end of observation}. Additionally, we utilize the open-source implementation of three additional state-of-the-art methods (denoted by $\dagger$) and evaluate them on the TrajNet++ benchmark. Compared to these competing baselines, SGANv2 improves upon the Top-3 ADE/FDE metric by $\sim$ 10\% and the collision metric by $\sim$ 40\%.  

We perform collaborative sampling to refine trajectories that undergo collision in real world datasets. For each colliding prediction, we perform 5 refinement iterations with stepsize 0.01. We observe that this procedure reduces the collision rate by \textbf{$\sim$ 30\%} on both ETH/UCY and TrajNet++. The trained discriminator understands human social interactions, and provides feedback to the bad samples, and consequently helps to reduce collisions. The second row of Fig~\ref{fig:collab_sample_viz} illustrates a few real-world scenarios where collaborative sampling demonstrates the ability to refine generator predictions that undergo collisions. In conclusion, we observe that SGANv2 beats competitive baselines in generating socially-compliant trajectories without compromising on the distance-based metrics.


\subsection{Ablation: Interaction Modelling}
In Table~\ref{interaction}, we empirically demonstrate that modelling interactions is the key to reducing prediction collisions. We consider the performance of different variants of our proposed SGANv2 architecture based on the interaction modelling schemes within the generator and discriminator. It is apparent that modelling interaction within both the generator and discriminator is necessary to output safe multimodal trajectories.

\begin{table}[!t]
    \centering
    \resizebox{0.45\textwidth}{!}{\begin{tabular}{cccccc}
    \toprule
 \multicolumn{1}{c}{$\textbf{G}_{Pool}$} & \multicolumn{1}{c}{$\textbf{D}_{Pool}$} & \multicolumn{2}{c}{\textbf{TrajNet++ Synth}} & \multicolumn{2}{c}{\textbf{TrajNet++ Real}} \\ \rowcolor{gray!15}
 \; & \; & Top-3 & Col & Top-3 & Col
\\ \midrule
 \xmark     & \xmark     & 0.3 / 0.5 & 18.3 & 0.5 / 1.1 & 9.6  \\ \rowcolor{gray!15}
 \checkmark & \xmark     & 0.2 / 0.4 & 4.1  & 0.5 / 1.0 & 3.9  \\ 
 \checkmark & \checkmark & 0.2 / 0.4 & \textbf{2.9}  & 0.5 / 1.0 & \textbf{3.1}  \\ 
 \bottomrule
\end{tabular}}
\caption{Interaction modules of SGANv2. Errors reported are Top-3 ADE/FDE (in m) and collision (in \%). Modelling interactions greatly reduces collisions on TrajNet++.}
\label{interaction}
\end{table}

\begin{figure*}[t!]
\centering
\begin{subfigure}[t]{0.32\textwidth}
    \includegraphics[width=\textwidth]{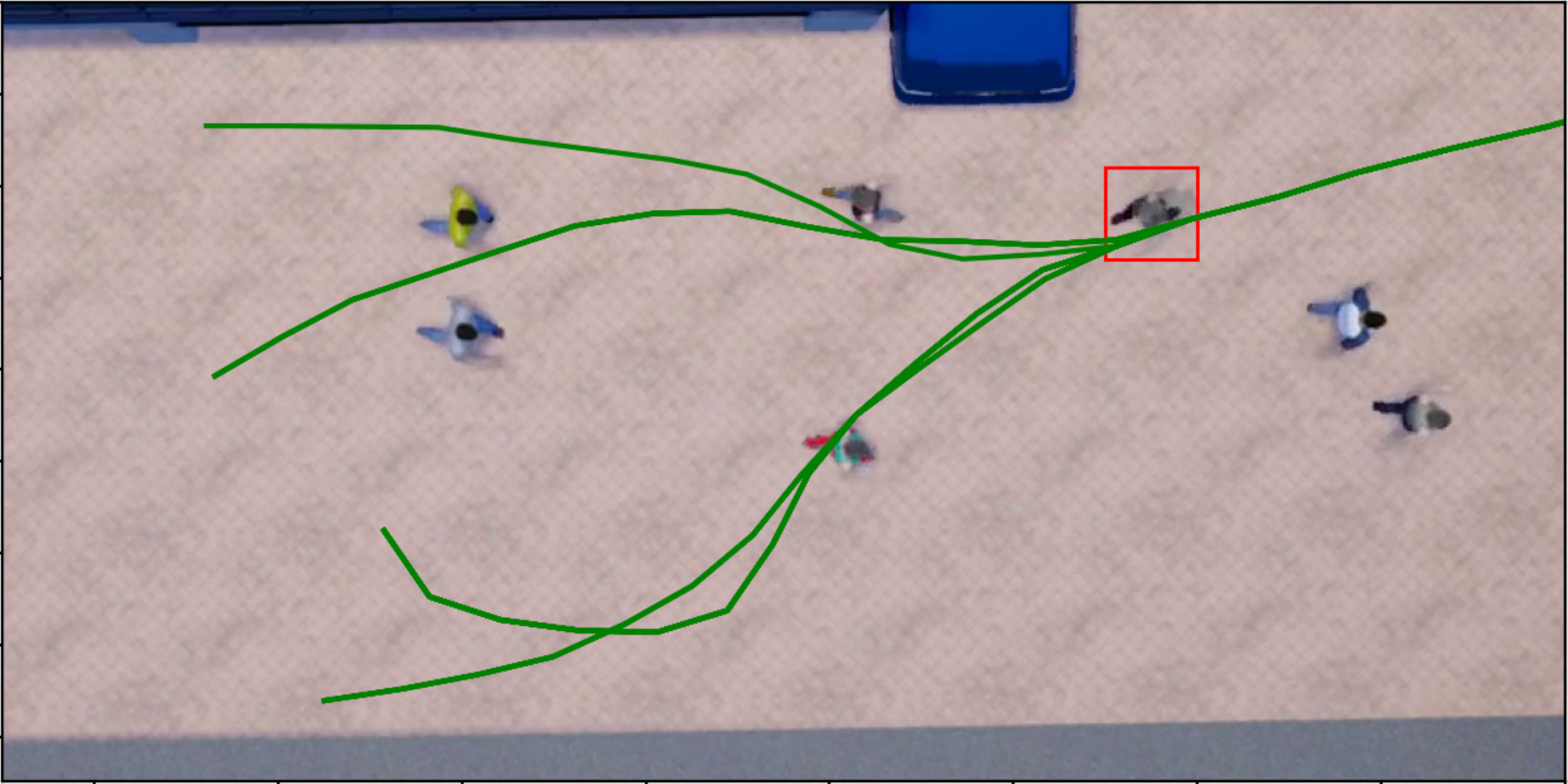}
    \caption{Ground Truth}
\end{subfigure}
\begin{subfigure}[t]{0.32\textwidth}
    \includegraphics[width=\textwidth]{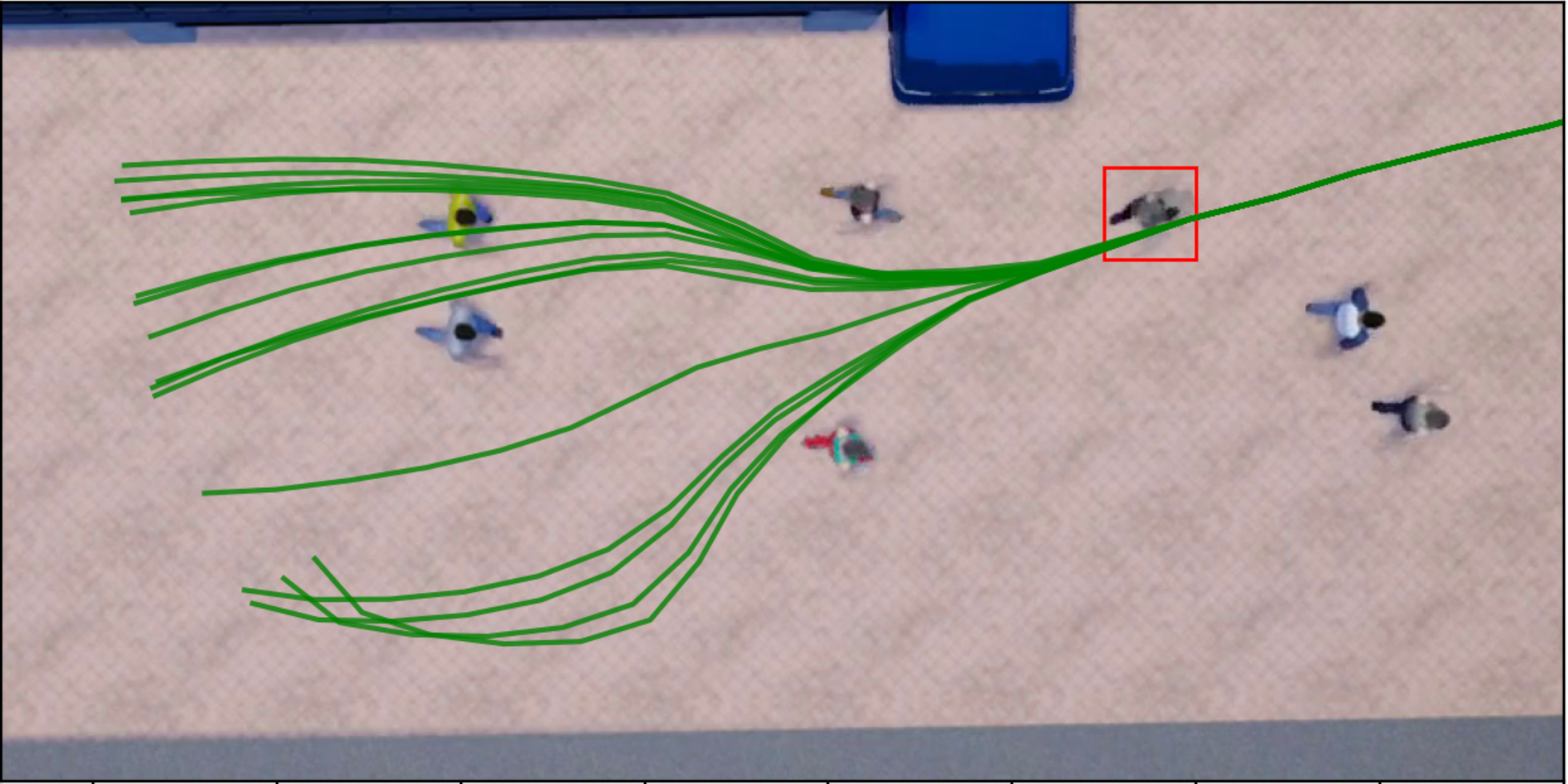}
    \caption{Variety Loss}
    \label{fig:var}

\end{subfigure}
\begin{subfigure}[t]{0.32\textwidth}
    \includegraphics[width=\textwidth]{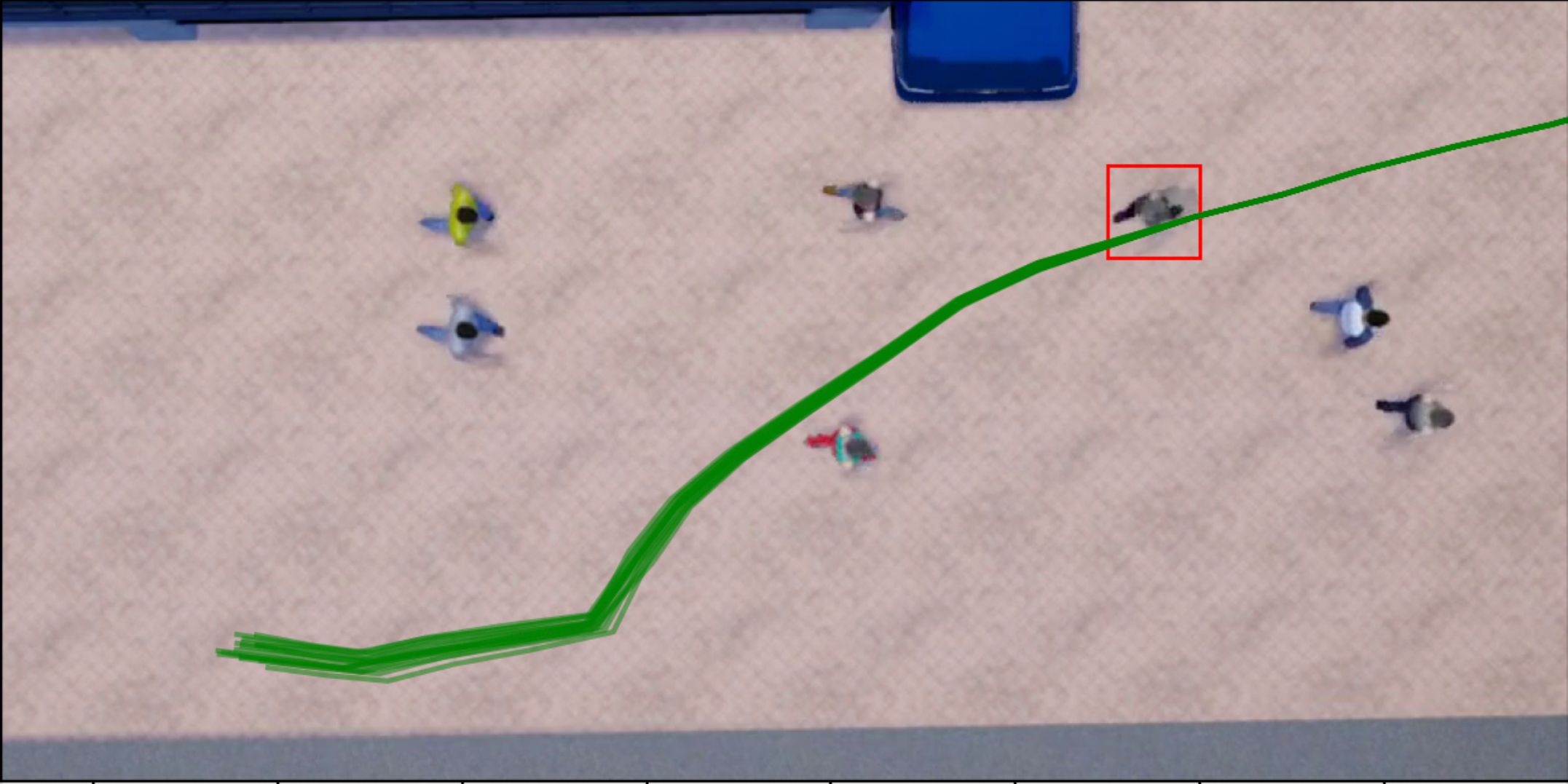}
    \caption{SGAN}
    \label{fig:sgan}

\end{subfigure} \\
\begin{subfigure}[t]{0.32\textwidth}
    \includegraphics[width=\textwidth]{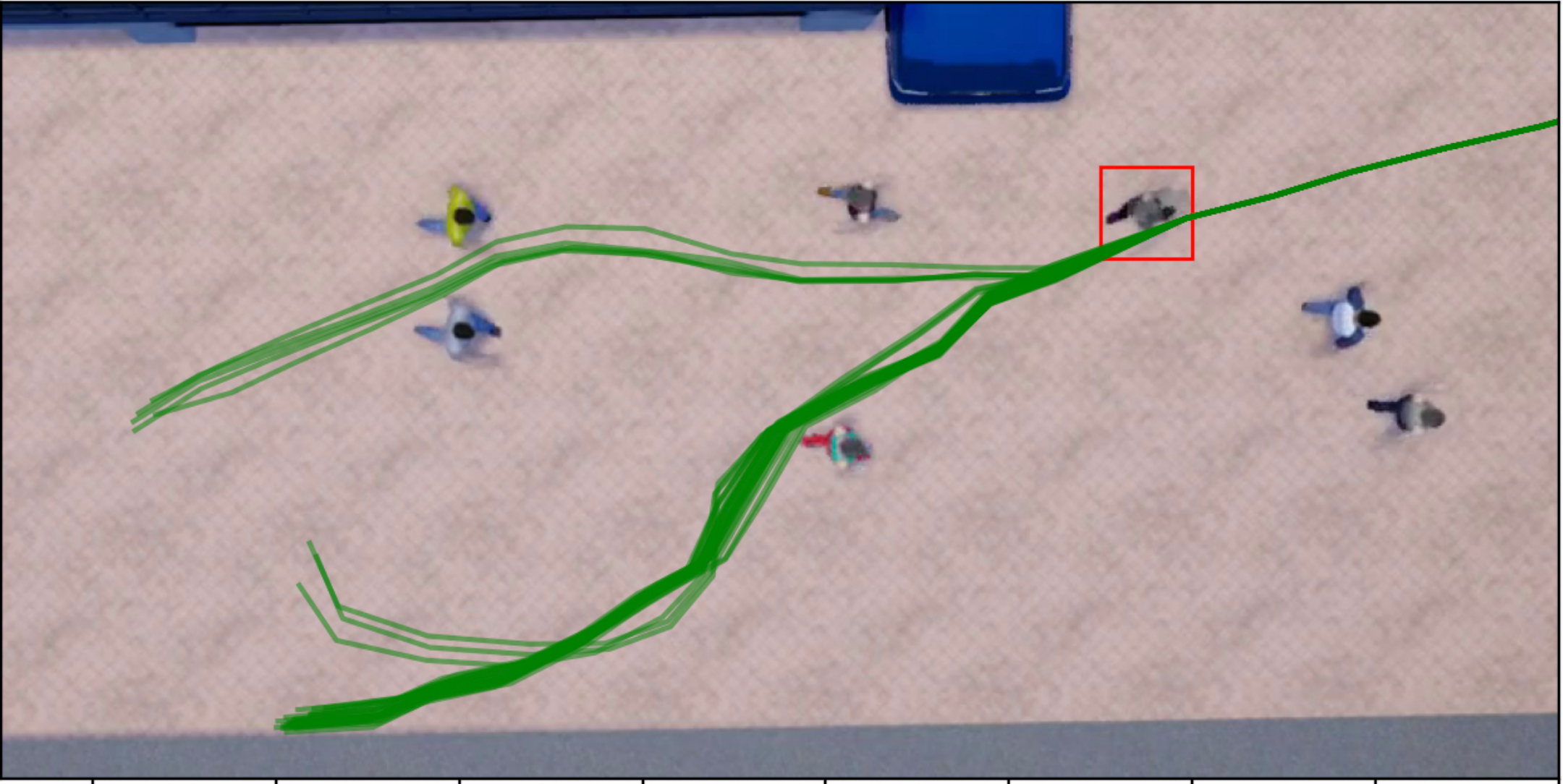}
    \caption{Social Ways}
    \label{fig:infoGan}

\end{subfigure}
\begin{subfigure}[t]{0.32\textwidth}
    \includegraphics[width=\textwidth]{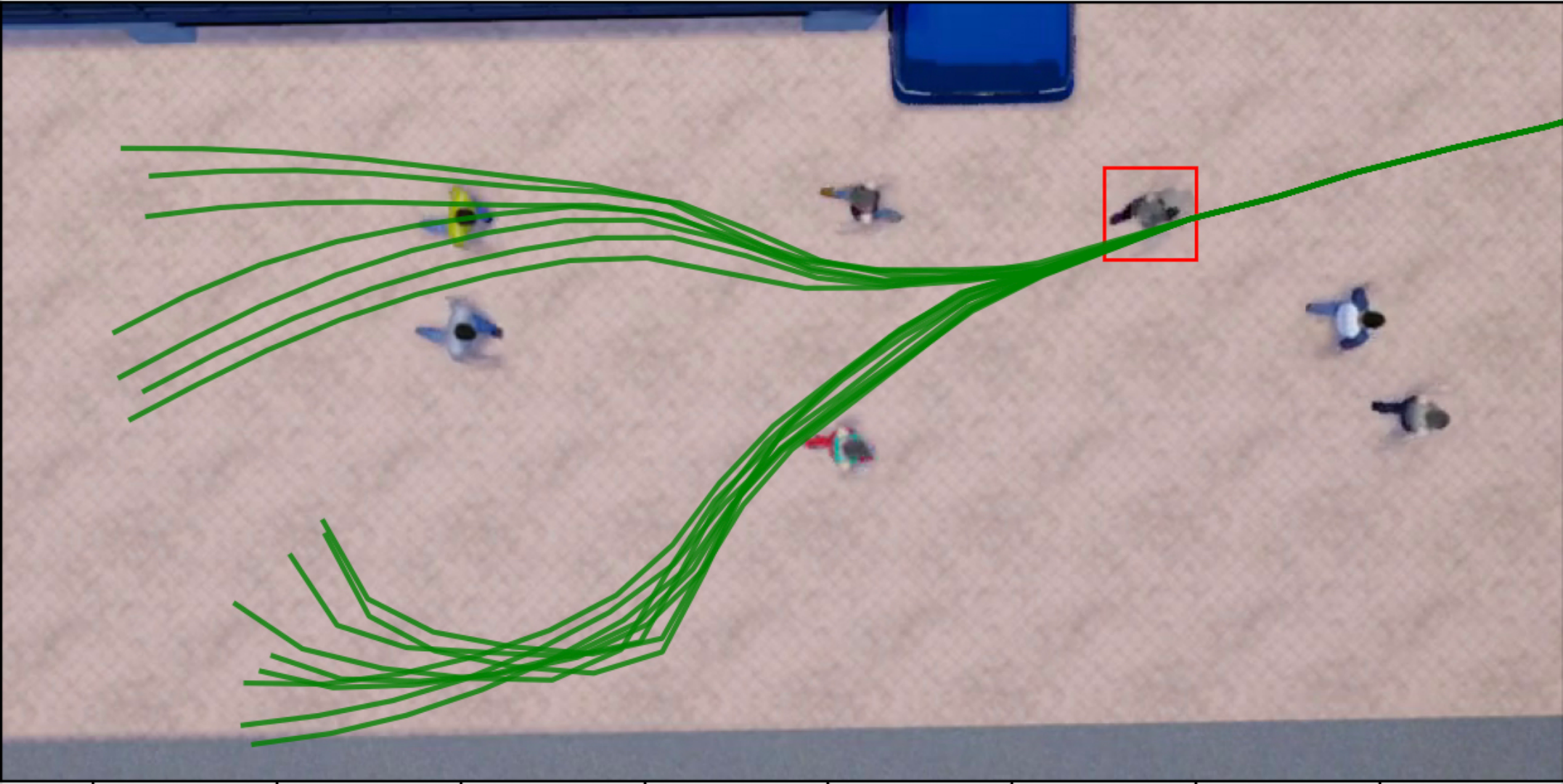}
    \caption{SGANv2 w/o CS}
    \label{fig:WGan}

\end{subfigure}
\begin{subfigure}[t]{0.32\textwidth}
    \includegraphics[width=\textwidth]{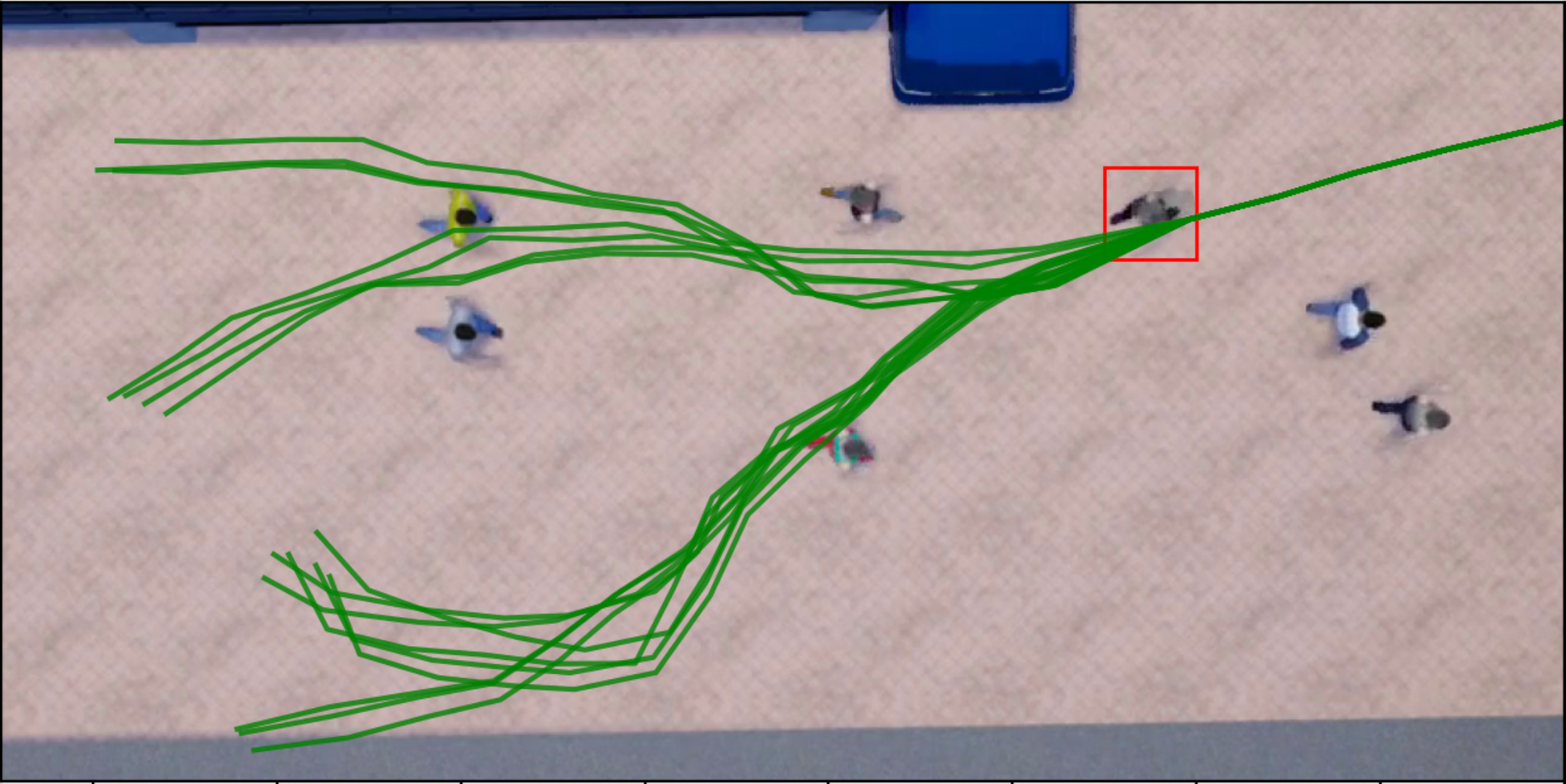}
    \caption{SGANv2}
    \label{fig:WGan_collab}

\end{subfigure}
\caption{Qualitative illustration of effectiveness of collaborative sampling on Forking Paths \cite{liang2020garden}. (b) Training models using variety loss \cite{Rupprecht2016LearningIA} leads to uniform output distribution. (c) Training using SGAN objective \cite{Gupta2018SocialGS} leads to mode collapse while (d) InfoGAN \cite{Amirian2019SocialWL} helps to mitigate the mode collapse issue. (e) SGANv2 helps to cover all the modes, and in combination with (f) collaborative sampling, we can successfully recover all modes with high accuracy.}
\label{fig:collab_fig}
\end{figure*}

\subsection{Multimodal Analysis}
In this final experiment, we demonstrate the potential of collaborative sampling to prevent mode collapse in trajectory generation. We utilize the sample scene `Zara01' from the Forking Paths dataset. We choose this scene as the multimodal futures of the `Zara01' scene is only affected by social interactions, and not physical obstacles. It forms the ideal test ground to check the multimodal performance of forecasting models. In this experiment, we observe the trajectories for 8 times steps (3.2 seconds) and show prediction results for 13 (5.2 seconds) time steps. 

Fig.~\ref{fig:collab_fig} qualitatively illustrates the performance of a GAN model trained using variety loss \cite{Rupprecht2016LearningIA, Gupta2018SocialGS} and other GAN objectives on the chosen scene. As there are 4 dominant modes in the scene, we chose $k=4$ for the variety loss. The model trained using variety loss (Fig.~\ref{fig:var}) ends up learning a uniform distribution, \textit{i.e.}, high diversity and low quality, as there is no penalty on the bad samples during training. Variety loss only penalizes the sample closest to the ground-truth. SGAN training \cite{Gupta2018SocialGS} (Fig.~\ref{fig:sgan}) results in mode collapse, \textit{i.e.}, low diversity and high quality as standard GAN training is highly unstable. Social Ways \cite{Amirian2019SocialWL} proposed infoGAN objective \cite{chen_infogan:_2016} to mitigate the mode collapse issue. The InfoGAN improves upon SGAN, however, it still fails to cover all the modes (Fig.~\ref{fig:infoGan}).

Empirically, we found that training SGANv2 with the gradient penalty objective (Fig.~\ref{fig:WGan}), proposed in \cite{arjovsky_wasserstein_2017}, provides a better mode coverage compared to InfoGAN, but the resulting distribution is still not accurate.  As shown in Fig.~\ref{fig:WGan_collab}, our proposed collaborative sampling at test-time helps to improve the accuracy of the SGANv2 predictions, recovering modes with low coverage. The trained discriminator guides the generated samples to these modes. Thus, we see that collaborative sampling is not only effective in refining trajectories at test time, but also can help to prevent mode collapse.

\subsection{Key attributes}

We now analyze the performance of the key SGANv2 design choices in the TrajNet++ synthetic setup. In the synthetic setup, we have access to the goals of each agent, allowing us to calculate Distance-to-Goal (Dist2Goal) \cite{Ma2017ForecastingID}, defined as the L2 distance between the predicted final destination and the goal of the agent. 

\begin{figure}[t]
\centering
\begin{subfigure}[t]{0.40\textwidth}
    \includegraphics[width=\textwidth]{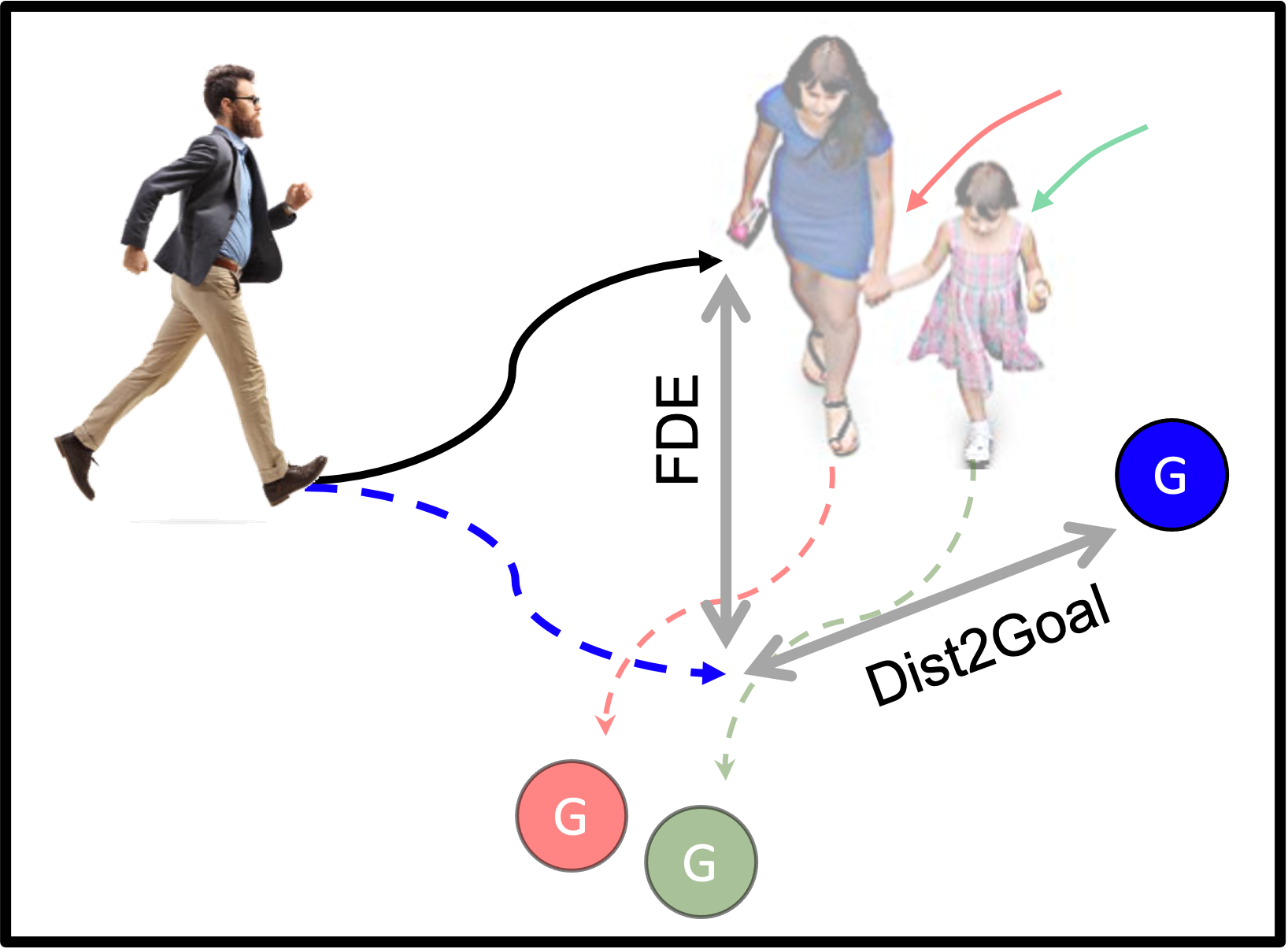}
\end{subfigure}
\caption[Illustration of the Distance-To-Goal metric.]{Illustration of difference between Dist2Goal and Final Displacement Error (FDE). FDE is the distance between ground-truth position and predicted position at end of prediction period. Dist2Goal is the distance between predicted position and the final position of agent in the entire dataset.}
\label{fig:dist2goal_main}
\end{figure}

\textbf{Rationale behind Distance to Goal:} It is possible that the generator predicts a socially-acceptable mode that \textit{does not correspond} to the ground-truth mode (see Fig.~\ref{fig:dist2goal_main}). If we calculate the ADE/FDE with respect to the ground-truth for such a predicted mode (that differs from ground-truth), the numbers will be high, \textit{misleading} us to incorrectly conclude that the generator did not learn the underlying task of trajectory forecasting. However, if the predicted destination is close to the goal of the agent, then one can assert that a different but socially acceptable mode has been predicted. The Col metric will help to validate that no collisions take place. Thus, Dist2Goal in combination with the Col metric helps to validate that a predicted mode is socially plausible.

Table~\ref{tab:traj_synth_ablat_main} quantifies the performance of various GAN architectures trained \textit{without} variety loss \cite{Rupprecht2016LearningIA}. SGAN \cite{Gupta2018SocialGS} performs the worst on the Col metric as the discriminator does not perform any interaction modelling, thereby not possessing the ability to learn the concept of collision avoidance. Only if the discriminator learns the collision avoidance property, can we expect it to teach the generator to output collision-free trajectories.
The global discriminator of S-BiGAT \cite{Kosaraju2019SocialBiGATMT} performs spatial interaction modelling \textit{only once}, at the end of prediction. Thus, the global discriminator is able to reason about interactions spatially but cannot model the temporal evolution of the same. SGANv2 equipped with spatio-temporal interaction modelling results in \textbf{near-zero} prediction collision. It is apparent that spatio-temporal interaction modelling within the discriminator plays a significant role in teaching the generator the concept of collision avoidance.

We now justify the design choices of sequence modelling within the discriminator using the \textit{Dist2Goal} metric. We compare an additional design of our proposed SGANv2 architecture: SGANv2-L, an SGANv2 with an LSTM discriminator. SGANv2-L trained using LSTM discriminator shows stopping behavior, indicated by the high Dist2Goal value in the test set. In other words, SGANv2-L outputs collision-free trajectories but the predictions fail to move towards the goal of the primary agent. In comparison, SGANv2 is able to output collision-free trajectories with a lower Dist2Goal (almost matching the ground-truth Dist2Goal value of $8.6m$). In conclusion, SGANv2 is able to output socially acceptable trajectories when compared to other GAN-based designs.
    
\begin{table}[t]
\centering
\resizebox{0.48\textwidth}{!}{
\begin{tabular}{ lcccc } 
\toprule
Model & \makecell{Spatio-temporal \\ Interaction Modelling \\ in Discriminator} & \makecell{Discriminator \\ Design} & Col & Dist2Goal\\ 
\midrule
 Ground-truth & -- & -- & 0.0 & 8.6 \\  \rowcolor{gray!15}
 SGAN \cite{Gupta2018SocialGS}  & \xmark & LSTM  & 24.9 & 8.9  \\
 S-BiGAT \cite{Kosaraju2019SocialBiGATMT}  & \xmark & LSTM & 8.4 & 8.9\\  \rowcolor{gray!15}
 SGANv2-L & \checkmark & LSTM  & 0.8 & 8.8\\
 SGANv2 & \checkmark  & Transformer & \textbf{0.2} & \textbf{8.6} \\
 \bottomrule
\end{tabular}}
\caption[Quantitative evaluation of various SGANv2 design choices on TrajNet++ synthetic dataset.]{Quantitative evaluation of various GAN architectural designs on TrajNet++ synthetic dataset. Col in \% and Dist2Goal in meters. SGANv2 learns to successfully predict socially acceptable outputs evidenced by lower collisions and Dist2Goal.}
\label{tab:traj_synth_ablat_main}
\end{table}

\subsection{Computational Time.}
Speed is crucial for a method to be used in a real world setting like autonomous vehicles where you need accurate predictions about pedestrian behavior. We provide the computational time at inference for our method against baseline unimodal LSTMs with and without interaction modelling. All the run times have been benchmarked on a single NVIDIA 2080 Ti GPU. We provide the run time per scene (averaged over all the scenes in the TrajNet++ real world benchmark).

\begin{table}[h]
    \centering
    \begin{tabular}{|c|c|c|c|c|}
    \hline
                & LSTM & D-LSTM & SGANv2 w/o CS & SGANv2 \\
    \hline
        Time    &     10ms    &   22ms  &  22ms  & 77ms \\
   \hline
    \end{tabular}
    \caption{Computational time comparison at inference per scene for various forecasting designs. The additional computational time for SGANv2 corresponds to the sample refinement process that occurs for five iterations.}
    \label{tab:compute}
\end{table}


The runtimes of D-LSTM and SGANv2 without collaborative sampling are similar as the multiple future predictions in the latter case can be generated in parallel, albeit at the cost of additional memory complexity. The relatively higher computational time of collaborative sampling corresponds to the sample refinement process based on the gradients from the discriminator. Nevertheless, the absolute computational time of collaborative sampling (77ms per scene) is suitable for real-time applications like autonomous systems.

\subsection{Implementation details}
The generator and the discriminator have their own spatial interaction embedding modules (SIM). Each pedestrian has his/her encoder and decoder.

\paragraph{Synthetic experiments.}\label{synth_param} The velocity of each pedestrian is embedded into a 16-dimensional vector. The hidden-state dimension of the encoder LSTM and decoder LSTM of the generator is 64. The dimension of the interaction vector of both the generator and discriminator is fixed to 64.  We utilize Directional-Grid \cite{Kothari2020HumanTF} interaction module with a grid of size $12 \times 12$ and a resolution of $0.6$ meters. For the LSTM discriminator, the hidden-state dimension is set to 64. For the transformer-based discriminator, we stack N=4 encoder layers together. The dimension of query vector, key vector and value vector is fixed to 64. The dimension of the feedforward hidden layer within each encoder layer is set to 64. We train using ADAM optimizer \cite{Kingma2015AdamAM} with a learning rate of 0.0003 for the generator and 0.001 for the discriminator for 50 epochs. The ratio of generator steps to discriminator steps for LSTM discriminator and transformer-based discriminator is 2:1. For synthetic data experiment, we have access to the goals of each pedestrian. The direction to the goal is embedded into a 16-dimensional vector. The batch size is fixed to 32. 

\paragraph{Real-world experiments.}\label{real_param}
The velocity of each pedestrian is embedded into a 32-dimensional vector. The hidden-state dimension of the encoder LSTM and decoder LSTM of the generator is 128. The dimension of the interaction vector of both the generator and discriminator is fixed to 256. We utilize Directional-Grid \cite{Kothari2020HumanTF} interaction module with a grid of size $12 \times 12$ and a resolution of $0.6$ meters. For the LSTM discriminator, the hidden-state dimension is set to 128. The ratio of generator steps to discriminator steps is 2:1. For the transformer-based discriminator, we stack N=2 encoder layers together (see Fig. 2 of main text). The dimension of query vector, key vector and value vector is fixed to 128. The dimension of the feedforward hidden layer within each encoder layer is set to 1024. We train using ADAM optimizer \cite{Kingma2015AdamAM} with a learning rate of 0.001 for both the generator and the discriminator for 25 epochs with a learning rate scheduler of step-size 10. The batch size is fixed to 32. The weight of variety loss is set to 0.2.

\paragraph{Multimodal Analysis.} The velocity of each pedestrian is embedded into a 16-dimensional vector. The hidden-state dimension of the encoder LSTM and decoder LSTM of the generator is 32. We train using ADAM optimizer \cite{Kingma2015AdamAM} with a learning rate of 0.0003 for the generator and 0.001 for the discriminator.

\section{Conclusion}
We presented SGANv2, an improved SGAN architecture equipped with two crucial architectural changes in order to output safety-compliant trajectories. First, SGANv2 incorporates spatio-temporal interaction modelling that can help to understand the subtle nuances of human interactions. Secondly, the transformer-based discriminator better guides the generator learning process. Furthermore, the collaborative sampling strategy helps leverage the trained discriminator during test-time to identify and refine the socially-unacceptable trajectories output by the generator. We empirically demonstrated the strength of SGANv2 to reduce the model collisions without comprising the distance-based metrics. We additionally highlighted the potential of collaborative sampling to overcome mode collapse in a challenging multimodal scenario. 

Our work aims at expanding the current horizon of trajectory forecasting models for real-world applications where humans' lives are at risk, such as social robots or autonomous vehicles. Accuracy, safety, and robustness are all mandatory keywords. Over the past years, researchers have focused their evaluation on distance-based metrics. Yet, if we compare the methods on the safety-critical ``collision" metric, we observe a difference in performance above 50\%. Hence, we believe that one should focus more on this metric and develop methods that aim for zero collisions.

\section*{Acknowledgement}
This work was supported by the Honda R\&D Co. Ltd and EPFL. We also thank VITA members and reviewers for their valuable comments.



%

\ifCLASSOPTIONcaptionsoff
  \newpage
\fi

\bibliographystyle{IEEEtran}
\bibliography{main.bib}{}

\begin{IEEEbiography}[{\includegraphics[width=1in,height=1.25in,clip,keepaspectratio]{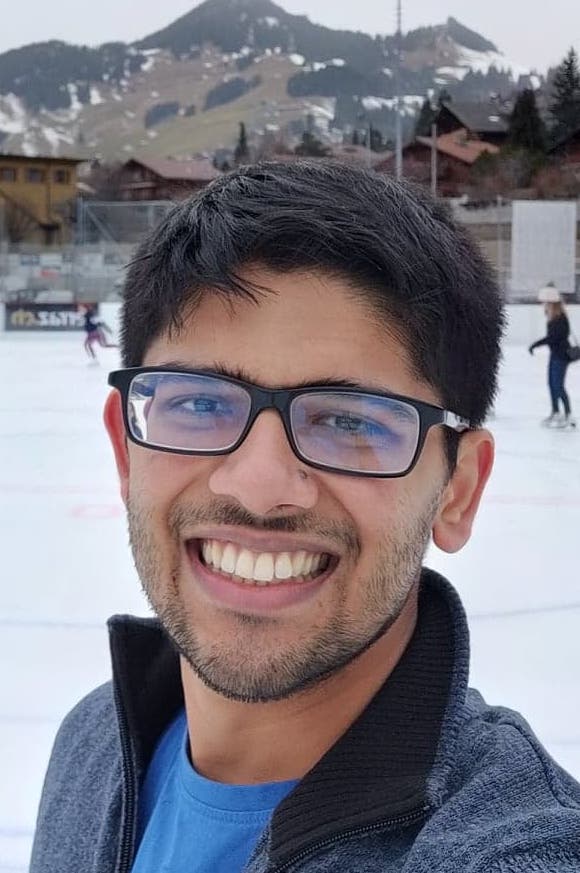}}]{Parth Kothari} is a fourth year doctoral student at Visual Intelligence for Transportation (VITA) Lab at EPFL, Switzerland. He received his Bachelors degree with Honours in Electrical Engineering from the Indian Institute of Technology, Bombay in 2018. His research interests include trajectory forecasting, modelling crowd behaviors and deep learning.
\end{IEEEbiography}



\begin{IEEEbiography}[{\includegraphics[width=1in,height=1.25in,clip,keepaspectratio]{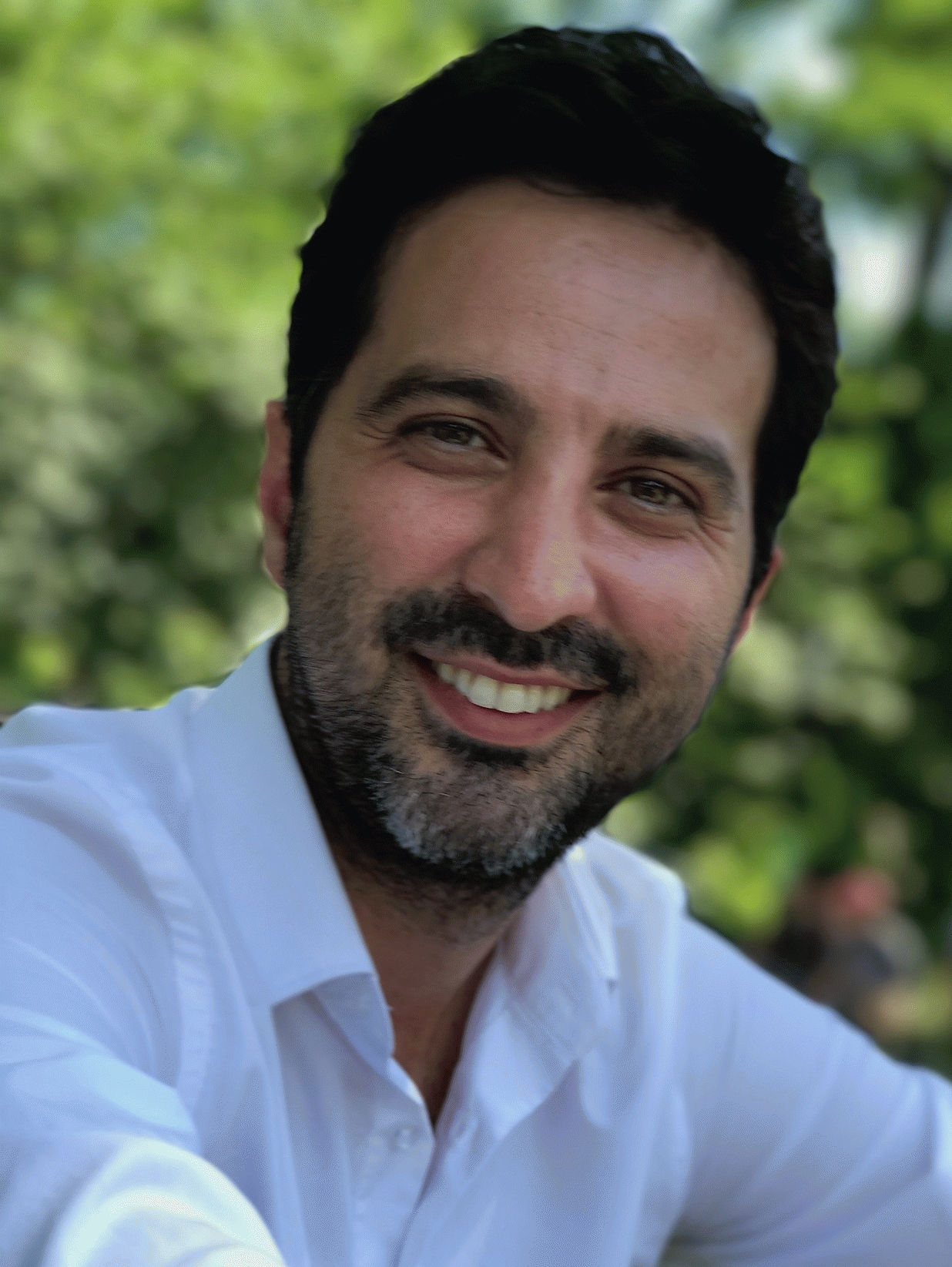}}]{Alexandre Alahi} is an Assistant Professor at EPFL. He spent five years at Stanford University as a Post-doc and Research Scientist after obtaining his Ph.D. from EPFL. His research enables machines to perceive the world and make decisions in the context of transportation problems and smart environments. He has worked on the theoretical challenges and practical applications of socially-aware Artificial Intelligence, \textit{i.e.}, systems equipped with perception and social intelligence. He was awarded the Swiss NSF early and advanced researcher grants for his work on predicting human social behavior. Alexandre has also co-founded multiple startups such as Visiosafe, and won several startup competitions. He was elected as one of the Top 20 Swiss Venture leaders in 2010.
\end{IEEEbiography}

\end{document}